\documentclass{article}

\PassOptionsToPackage{numbers, sort&compress, comma, square}{natbib}
\usepackage[preprint]{neurips_2021}

\usepackage[utf8]{inputenc} 
\usepackage[T1]{fontenc}    
\usepackage{hyperref}       
\usepackage{url}            
\usepackage{booktabs}       
\usepackage{amsfonts}       
\usepackage{nicefrac}       
\usepackage{microtype}      
\usepackage[dvipsnames]{xcolor}
\usepackage{amsmath}
\usepackage{xspace}
\usepackage{tabularx}
\usepackage{wrapfig,lipsum}
\usepackage{graphicx} 
\usepackage{multirow}
\usepackage{CJK}

\graphicspath{ {./images/} }

\newcommand{\ttbf}[1]{\texttt{\textbf{#1}}}
\newcommand{\mf}[0]{\ttbf{NRETM}\xspace}
\newcolumntype{Y}{>{\centering\arraybackslash}X}
\newcolumntype{L}{>{\raggedright\arraybackslash}X}

\title{Neural Rule-Execution Tracking Machine For Transformer-Based Text Generation}

\author{%
  Yufei Wang
  \thanks{Work done during the internship at Microsoft STCA.}
  \\
  Macquarie University\\
  {\small\texttt{yufei.wang@students.mq.edu.au}} \\
  \And
  Can Xu \\
  Microsoft STCA NLP \\
  {\small \texttt{caxu@microsoft.com}} \\
  \And
  Huang Hu \\
  Microsoft STCA NLP \\
  {\small\texttt{huahu@microsoft.com}} \\
  \And
  Chongyang Tao \\
  Microsoft STCA NLP \\
  {\small\texttt{chongyang.tao@microsoft.com}} \\
  \And
  Stephen Wan \\
  CSIRO Data61 \\
  {\small\texttt{stephen.wan@data61.csiro.au}} \\
  \And
  Mark Dras \\
  Macquarie University \\
  {\small\texttt{mark.dras@mq.edu.au}} \\
  \And
  Mark Johnson \\
  Macquarie University \\
  {\small\texttt{mark.johnson@mq.edu.au}} \\
  \And
  Daxin Jiang\thanks{Corresponding author: Daxin Jiang (djiang@microsoft.com).} \\
  Microsoft STCA NLP \\
  {\small\texttt{djiang@microsoft.com}} \\
}

\begin{document}

\maketitle

\begin{abstract}
Sequence-to-Sequence (Seq2Seq) neural text generation models, especially the pre-trained ones (e.g., BART and T5), have exhibited compelling performance on various natural language generation tasks. However, the black-box nature of these models limits their application in tasks where specific rules (e.g., controllable constraints, prior knowledge) need to be executed. Previous works either design specific model structures (e.g., Copy Mechanism corresponding to the rule ``the generated output should include certain words in the source input'') or implement specialized inference algorithms (e.g., Constrained Beam Search) to execute particular rules through the text generation. These methods require the careful design case-by-case and are difficult to support multiple rules concurrently. In this paper, we propose a novel module named Neural Rule-Execution Tracking Machine, i.e., \texttt{NRETM}, that can be equipped into various transformer-based generators to leverage multiple rules simultaneously to guide the neural generation model for superior generation performance in an unified and scalable way. Extensive experiments on several benchmarks verify the effectiveness of our proposed model in both controllable and general text generation tasks.
\end{abstract}

\section{Introduction}
Transformer-based neural language models (LMs), such as GPT/BART~\cite{radford2019language,brown2020language,lewis-etal-2020-bart}, have led a wave of new trends in natural language generation, producing texts of prominent quality. Such LMs are usually trained roughly on huge amounts of text corpora to maximize the likelihoods of predicting next tokens. Despite their success in varieties of NLP tasks, we argue that the black-box nature of these models leads to inefficiently learning to follow constraints and incorporating prior knowledge.

In controllable text generation, most relevant studies~\cite{keskar2019ctrl,ziegler2019fine,dathathri2019plug} focus on controlling high-level text attributes (e.g., topic, sentiment) or simply keyword/phrase. More complex fine-grained control constraints such as ``generate a text with `apple' in the second sentence which has 15 words and `orange' or `oranges' in the sixth sentence'' are less explored. A very recent work \cite{lu2020neurologic} reveals that large-scale LMs do not learn to obey the underlying constraints reliably, even in a quite simple constrained generation task (cover all the given keywords without hallucinating new ones). It is conceivable that how such LMs will behave when expected to follow the propositional control constraints mentioned above. In general text generation, existing works on various tasks reveal the benefit of incorporating task-specific prior knowledge: machine translation~\cite{tu2016modeling} (e.g., the coverage constraint that each source phrase should be translated into exactly one target phrase), text summarization~\cite{yang2020ted} (e.g., the lead bias: front loading the most salient information), dialogue generation~\cite{gu-etal-2016-incorporating} (e.g., humans tend to repeat entity names or even long phrases in conversation). However, they either need designing specific model architectures (e.g., Coverage Mechanism and Copy Mechanism) or devising well-designed learning objectives (e.g., GSG~\cite{zhang2020pegasus}). These methods require careful design case-by-case and are difficult to combine multiple arbitrary prior knowledge simultaneously.

The dilemma of the above two research lines motivates us to explore unifying the constrained generation and prior knowledge integration into one single generation framework that could effectively execute multiple rules (defined in predicate logic form) simultaneously. Thus, we propose to equip the transformer-based sequence-to-sequence architecture with a novel module named Neural Rule-Execution Tracking Machine (\mf)~\footnote{We will make the code public available later to facilitate reproducing the results}, which effectively enforces the satisfaction of given constraints by dynamically monitoring the expression progress of each constraint during the decoding process. We leverage executable programs as the logic checking operator, thus these constraints can be any predicate logic formula and a variety of predicate functions can be defined to support flexible expansion. \mf consists of a state matrix and a logical tracker where the former records status of constraint expression and the latter update the state matrix according to the latest decoding result. The state representations aggregated from the state matrix are injected into the decoder as the relative positions between encoder output and current decoder input in the cross-attention module. Our approach reconciles symbolic computing (that has precise logic and numerical calculation capabilities) with neural language generation (that has an exceptional ability of wording and phrasing), which results in both the accurate controllability and the superior generation performance.

We conduct experiments on three benchmarks: ROCStories~\cite{mostafazadeh-etal-2016-corpus} consisting five-sentence stories; Commonsense Generation~\cite{lin-etal-2020-commongen}, target explicitly test machines for the ability of generative commonsense reasoning; TED15 Zh-En~\cite{miculicich-etal-2018-document}, a common document-level machine translation benchmark. We design various of controllable and general text generation settings in these benchmarks and our \mf model significantly outperforms Seq2Seq pre-trained baselines.

Our contributions in this work are three-fold: (1) proposal of unifying constrained generation and prior knowledge incorporation in a highly customizable predicate logic controllable generation framework; (2) proposal of a neural rule-execution tracking machine that effectively guides the generation following any predicate logic constraints; and (3) empirical verification of the effectiveness of the proposed approach on three benchmarks. 
\section{Approach}
In this work, we formalize the execution of rules during text generation as generating sentences that conform to certain predicate logic constraints. We start from a definition of predicate logic constraint in our framework, an overview of the model, and then dive into details of each component.

\subsection{Predicate Logic Constraint}
We define a predicate $U$ with zero or more arguments. For example, $U(\textbf{a},\textbf{b})$ is an atomic formula with predicate $U$ and two arguments, here occupied by the variable $\textbf{a}$ and $\textbf{b}$. The predicate $U(\textbf{a},\textbf{b})$ is a boolean function asserting a relationship between variables $\textbf{a}$ and $\textbf{b}$, e.g., $\text{Copy}(\textbf{a},\textbf{b})$ means the occurrence of keyword $\textbf{a}$ in a sequence $\textbf{b}$, where $\text{Copy}$ is a specific kind of predicate and $\textbf{a}$ can be either unigram or multi-gram. In general, an argument is either a variable or a constant. Further, our method accepts predicate logic constraint $P_{c}$ in propositional logical formula like:
\[\big(U_1 \lor U_2 \cdots \lor U_i \big) \land \cdots \land \big(U_k \cdots \lor U_n \big) \]
where each $U_i$ represents a single positive or negative constraint, e.g., $U(\textbf{a}, \textbf{b})$ or $\neg U(\textbf{a}, \textbf{b})$, restricting whether $a$ and $b$ satisfy the relationship defined by $U$ or not, respectively. 
Our method seeks optimal generated sequence $\mathbf{y}$ conditioned on the input $\mathbf{x}$ in which $P_c$ are satisfied:
\begin{equation}
    \mathbf{y} {=} \arg\max_{\mathbf{y} \in \mathcal{Y}}P(\mathbf{y}|\mathbf{x}) \hspace{1em}\textrm{where}~ P_c{=}1
 \end{equation}
With the help of custom predicate logic functions, various control constraints and prior knowledge can be converted to the propositional logical formula, and thus handled by \mf. For example, the control constraint "generate a text with 'apple' in the second sentence which has 15 words and 'orange' or 'oranges' in the sixth sentence" can be transformed to "$\text{InSen}(\text{apple},2)\land\text{SenLen}(2,15)\land(\text{InSen}(\text{orange},6)\lor\text{InSen}(\text{oranges},6))$"; the prior knowledge "each source phrase should be translated into exactly one target phrase" can be transformed to "$\forall \> s_i \in \textbf{S}, \text{TranslatedOnce}(s_i)$", $s_i$ is the $i^{th}$ phrase in source. We define six kinds of predicates that used by \mf in this work, details please refer to the Supplementary Material A.2.

\begin{figure}[t]
\centering
\includegraphics[width=0.75\textwidth]{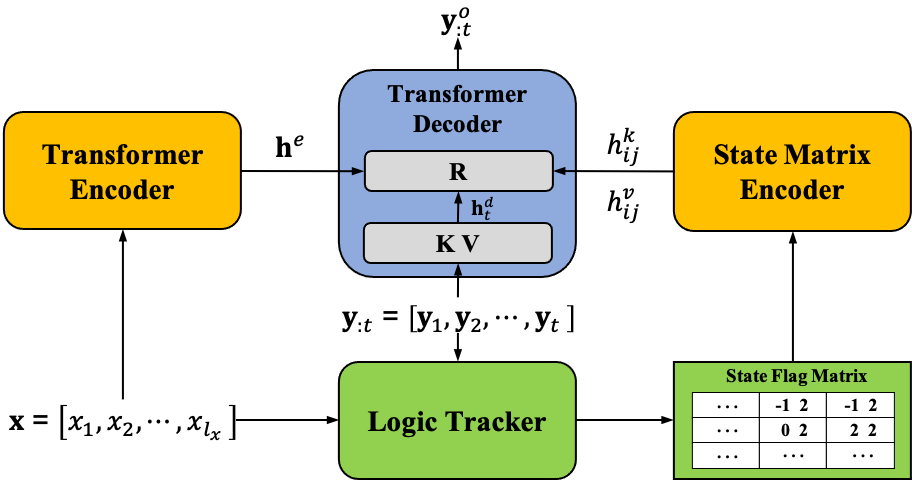}
\caption{An overview of our \mf model. Given logic constraints, input and current generated sequences, our \emph{Logic Tracker} maintains the current status of all logic constraints in State Flag Matrix, which is further processed by the State Matrix Encoder and fed into the Transformer-based text decoder to guide the text generation procedure.}
\label{overview}
\end{figure}

\subsection{Model Overview}
Figure~\ref{overview} illustrates the architecture of our neural rule-execution tracking machine (\mf). \mf is able to be equipped into any sequence-to-sequence transformer-based text generator, for the sake of simplicity, we take T5 for an example. Before output decoding, the input tokens are fed into transformer encoder, and the state matrix is initialized by the initial value of all predicates in constraints. Then, during response decoding, the state matrix tracks the expression of each predicate and gets updated by the Logic Tracker. The Logic Tracker manages the process of decoding at each step by checking the logical satisfaction of each predicate, and updating the status and corresponding intermediate value of each predicate into the state matrix. Based on the information provided by the state matrix, and the generated word sequence, the decoder predicts the next word of the output.

\subsection{Constrained Text Generation}
In the sequence-to-sequence constrained text generation tasks, we are given encoder inputs $\mathbf{x}=[x_1, \dots,x_{l_m},\dots, x_{l_x}] \in \mathcal{X}$, where $[x_{l_{m+1}},\dots, x_{l_x}]$ corresponds to predicate logic constraints that should be satisfied in the generated outputs. At generation step $t$, the decoder take $\mathbf{y}_{:t}=[y_1, \cdots, y_{t}] \in \mathcal{Y}$ as input and generate $\mathbf{y}_{t+1}$. 

\subsection{Neural Rule-Execution Tracking Machine}
The \mf dynamically controls text generation according to the given constraints via the cooperation of the state matrix and the Logic Tracker. It informs the decoder at the first time to what extend the constraints have been expressed. For the example shown in~\ref{overview}, the dynamic control strategy is more reasonable than static strategies such as concatenating the logic constraints with the input text to the encoder. This is because if the whole constraints are composed of five predicates and two predicates have been satisfied during decoding, then the decoder needs to know which three predicates have not been met and the intermediate status of each unfulfilled predicate rather than always memorizing that the five predicates should be all satisfied.

\textbf{State Flag:} Suppose that the given constraint $\mathcal{M}$ consists of $n$ predicates $\mathcal{D}=\{U_i\}_{i=1}^n$. At generation step $t$, there is a set of state flags indicates the expression progress of each predicate in $\mathcal{M}$ up to this step (i.e., in the decoder input sequence $\mathbf{y}_{:t}$). Formally, they can be defined as  $\mathrm{s}: \mathcal{X} \times \mathcal{Y} \rightarrow \{q_{it}\}_{i=1}^{l_x}$ where $|\mathrm{s}(\mathbf{x}, \mathbf{y}_{:t})|=l_x$, $q_{it}$ is the state flag in a string form. Specifically, state flag $\mathrm{s}(\mathbf{x}, \mathbf{y}_{:t})_i=q_{it}$ is for the input token $x_i$ in $\mathbf{x}$ at the decoding step $t$. As $q_{it}$ records the satisfaction of $\mathcal{D}$, $q_{it}$ is the concatenation of $\{q_{it}^k\}_{k=1}^n$, where $q_{it}^k$ is the $i^{th}$ expression state string of $U_k$ returned by Logic Tracker.

\textbf{Logic Tracker:} Suppose that \mf consist of $l$ unique predicates $\{\mathcal{U}_i\}_{i=1}^l$, then the Logic Tracker $\mathcal{L}$ is made up of $l$ logical operator $\{\mathcal{O}_i\}_{i=1}^l$, where $\mathcal{O}_i(\cdot,\cdot)$ is an executable program corresponding to $\mathcal{U}_i$, which takes the arguments of $\mathcal{U}_i$ and a default argument $\mathbf{y}_{:t}$ as inputs, and outputs expression state strings. Specifically, at step $t$ given a predicate $U_k$, $\mathcal{L}.\mathcal{O}_{\bar{k}}(U_k,\mathbf{y}_{:t}) = \{q_{it}^k\}_{i=1}^{l_x}$, where $q_{it}^k$ is the concatenation of default state status $\hat{q}_{it}^k$ and additional intermediate values $\bar{q}_{it}^k$ (it is optional, maybe empty). $\hat{q}_{it}^k$ is formulated as:
\begin{equation}
\hat{q}_{it}^k=\left\{
\begin{array}{rcl}
\text{N} & & {\text{variables of}~U_k~\text{do not contain}~x_i}  \\
0 & & {U_k~\text{is not satisfied}}  \\
1 & & {U_k~\text{is in progress}}\\
2 & & {U_k~\text{is satisfied}} \end{array} \right.
\label{mfdefine}
\end{equation}
The values 0 and 2 represent the status of predicate constraint satisfaction. At the start, for all tokens the value of default state status is set to 0. Once $\mathbf{y}_{:t}$ satisfies the constraints, value of the corresponding $\hat{q}_{it}^k$ is updated to 2. Value 1 is a state customized by each logical operator indicates that at current step t the predicate is not satisfied but triggered the beginning condition. $\bar{q}_{it}^k$ is another value that customized by each logical operator for better tracking, for example in the predicate $\text{Len}(\mathbf{y}, 20)$ (means the generated output should have a length of 20), $\bar{q}_{it}^k$ could be 20 minus the current length of $\mathbf{y}$ difference informing decoder how many words are left to satisfy the constraint. Each variable in $U_k$ takes one form from the following set $\{\mathbf{x}_{i:j}, \mathbf{y}_{i:j}, \mathbf{x}^i, \mathbf{y}^i\}$, where $\mathbf{x}_{i:j}$ and $\mathbf{y}_{i:j}$ are the sub-sequences of $\mathbf{x}$ and $\mathbf{y}$ from $i^{th}$ token to $j^{th}$ token respectively, $\mathbf{x}^i$ and $\mathbf{y}^i$ refer to $i^{th}$ sentence in $\mathbf{x}$ and $\mathbf{y}$ respectively. In a special case (i.e., variables in $U_k$ only have tokens from $\mathbf{y}$), $\hat{q}_{it}^k$ only have three states: 0,1,2.


\begin{figure}[t]
\centering
\includegraphics[width=\textwidth]{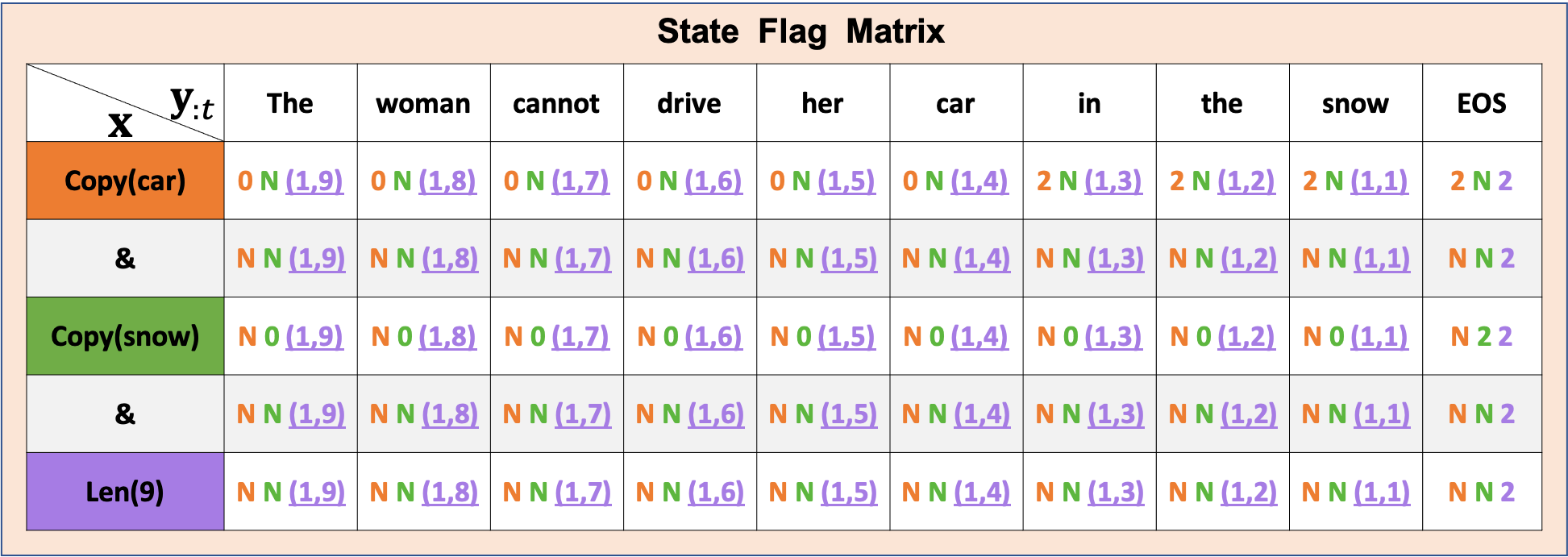}
\caption{A running example of our \mf model with three logic constraints (i.e., copy(car) \& copy(snow) \& len(9); copy car and snow in the output with 9 tokens). Each State Flag is a string that records the current status of all given logic constraints. Three tokens in each State Flag correspond to the constraints with the same color. Length constraint (``len(9)'') tracks intermediate values in {\color{RoyalPurple} purple color}. Copy constraints only track the final satisfaction status. All constraints are satisfied in the end.}
\vspace{-4mm}
\label{example}
\end{figure}

\textbf{Example:} In Figure~\ref{example}, we are given three logic constraints, \emph{a)} copy ``car''; \emph{b)} copy ``snow'' and \emph{c)} generate 9 tokens. Each State Flag has three values indicating the status of the three logic constraints. We assign ``N'' to the second constraints in the first row and the first constraints in the third row because \emph{a)} and \emph{b)} only associate with their corresponding words. Length control maintains a set of intermediate values (e.g., the residual length). While copy controls do not have intermediate values and they are updated from Value 0 to 2 only when the corresponding words (at step 7 and 10 in our example) are generated in the $\textbf{y}_{:t}$.

\textbf{State Matrix:} Given $\mathbf{x}$, $\mathbf{y}_{:t}$, we define the two-dimensional state matrix $\mathcal{S} \in \{q_{ij}\}^{l_x \times t}$ as follows:
\begin{align}
\mathcal{S} = [\mathrm{s}(\mathbf{x},\varepsilon); \mathrm{s}(\mathbf{x},\mathbf{y}_{:1}); \cdots; \mathrm{s}(\mathbf{x},\mathbf{y}_{:t})] 
\end{align}
where $\varepsilon$ is the empty string. During training, given $\mathbf{x}$ and the golden output $\mathbf{y}^{gt}$ (with $l_{gt}$ tokens), we can construct the ground-truth state matrix $\mathcal{S}^{gt} \in \{q_{ij}\}^{l_x \times l_{gt}}$ by utilizing each step of $\mathbf{y}^{gt}$ to get the state flag from the Logic Tracker $\mathcal{L}$. $\mathcal{S}^{gt}$ follows the same masking strategy as the decoder input tokens $\mathbf{y}_{:t}$. For the tokens whose corresponding predicate constraints having no alignment with $\mathbf{y}^{gt}$, their state flags are also assigned value N. During inference, we build the state matrix incrementally, starting from $\mathcal{S}^{\mathit{inf},0} = [\mathrm{s}(\mathbf{x},\varepsilon)] \in \{\text{N},0\}^{l_x \times 1}$. In step $t$, we add a new column $\mathrm{s}(\mathbf{x},\mathbf{y}_{:t})$ to $\mathcal{S}^{\mathit{inf},t-1} \in \{q_{ij}\}^{l_x \times (t-1)}$ and obtain the new state matrix $\mathcal{S}^{\mathit{inf},t} \in \{q_{ij}\}^{l_x \times t}$. 

\subsection{Integration with Transformer-Based Generator}
In a standard transformer-based encoder-decoder framework, given the input tokens $\{x_i\}_{i=1}^{l_{x}}$ and the generated tokens $\mathbf{y}_{:t}$, the probability of the next token $y_{t+1}$ can be calculated by:
\begin{align}
\mathbf{H}^e &= \mathrm{TransformerEncoder}(\{\mathbf{E}(x_i)\}_{i=1}^{l_{x}}) \\
\mathbf{H}_t^d &= \mathrm{KV}(\mathbf{W}_{q}^s \mathbf{E}(\mathbf{y}_{:t}), \mathbf{W}_{k}^s \mathbf{E}(\mathbf{y}_{:t}), \mathbf{W}_{v}^s \mathbf{E}(\mathbf{y}_{:t}))   \label{sa}\\
 o_{t+1} &= \mathrm{KV}(\mathbf{W}_{q}^c \mathbf{H}_{t}^{d}, \mathbf{W}_{k}^c \mathbf{H}^e, \mathbf{W}_{v}^c \mathbf{H}^e)  \label{ca} \\
p(y_{t+1}| x_{1:l_x},y_{1:t}) &= \mathrm{softmax}(\mathbf{W}_o \, o_{t+1})
\end{align}
where $\mathbf{E} \in \mathbb{R}^{d_e\times |V|}$ is the word embedding matrix with $d_e$ the dimension and $|V|$ the vocabulary size,  $o_{t+1} \in \mathbb{R}^{d_c}$ is the hidden state at step $t$ with $d_c$ the hidden size, and $\mathbf{W}_o \in \mathbb{R}^{|V|\times d_c}$ is a parameter matrix that maps the hidden state $o_{t+1}$ to a logit vector in the size of $|V|$, and $\mathrm{KV}$ is the standard key-value self-attention proposed in \citep{NIPS2017_3f5ee243}.

\textbf{State Matrix Encoder:} 
For each element $q_{ij}$ in the state matrix $\mathcal{S}$, we use the same tokenizer of the transformer-based encoder-decoder framework to obtain tokenized sequence $\{q_{ij}^k\}_{k=1}^{l_{ij}^q}$. Then we get the representation of $q_{ij}$ by feeding the tokenized sequence into a single-layer transformer-based encoder (same architecture as $\mathrm{TransformerEncoder}$) $\mathrm{ShallowEncoder}$:
\begin{align}
\mathbf{h}_{ij} &= \mathrm{ShallowEncoder}(\{\mathbf{E^q}(q_{ij}^t)\}_{t=1}^{l_{ij}^q}) \\
\bar{\mathbf{h}}_{ij} &= \mathrm{MeanPooling}(\mathbf{h}_{ij})
\end{align}
where $\mathbf{E^q} \in \mathbb{R}^{d_k\times |V|}$, which is frozen word embeddings initialized with the first $d_k$ dimension of $\mathbf{E}$, $\mathbf{h}_{ij} \in \mathbb{R}^{l_{ij}^q \times d}$, $\bar{\mathbf{h}}_{ij} \in \mathbb{R}^d$ and $d$ is the hidden size of $\mathrm{ShallowEncoder}$.

\textbf{State Matrix as Relative Position:} Inspired by~\cite{shaw-etal-2018-self} which incorporates token relative positions into the self-attention (Eq.~\ref{sa}) module, we propose to inject State Matrix as the ``relative positions'' between encoder output $\mathbf{H}^e$ and current decoder input $\mathbf{y}_{:t}$ in the cross-attention (Eq.~\ref{ca}) module. Following this approach, we linearly project each $\bar{\mathbf{h}}_{ij}$ into \emph{State Flag key} $\bar{\mathbf{h}}^k_{ij} = \mathbf{W}_{k}^f \cdot \bar{\mathbf{h}}_{ij} + \mathbf{b}_{k}^f$ and \emph{State Flag Value} $\bar{\mathbf{h}}^v_{ij} = \mathbf{W}_{v}^f \cdot \bar{\mathbf{h}}_{ij} + \mathbf{b}_{v}^f$. All transformer decoder layers share the same State Flags representations. Eq.~\ref{ca} is changed to:
\begin{align}
 o_{t+1} = \mathrm{R}(\mathbf{W}_{q}^c \mathbf{H}_{t}^{d}, \mathbf{W}_{k}^c \mathbf{H}^e, \mathbf{W}_{v}^c \mathbf{H}^e, \bar{\mathbf{h}}^k, \bar{\mathbf{h}}^v)
\end{align}
where $\bar{\mathbf{h}}^k$ and $\bar{\mathbf{h}}^v \in \mathbb{R}^{l_x \times t \times d}$ and $\mathit{R}$ is the Self-Attention function with relative position, defined as follows:
\begin{align}
\mathit{R}(\mathbf{q}, \mathbf{k}, \mathbf{v}, \mathbf{m}^k, \mathbf{m}^v)_j &= \sum_{i=1}^{l_x} \mathbf{a}_{i,j} (\mathbf{v}_i + \mathbf{m}_{i,j}^v) \\
\mathbf{a}_{*,j} &= \mathit{Softmax}(\mathbf{e}_{*,j}) \label{beforesoftmax} \\
\mathbf{e}_{i,j} &= \frac{\mathbf{q}_j (\mathbf{k}_i + \mathbf{m}^k_{i,j})^{T}}{\sqrt{d}}
\end{align}

\subsection{Why NRETM Could Satisfy Constraints}


Intuitively, we just make the neural generator aware of the expression state of each predicate logic constraint rather than forcing the model to execute these logical expressions explicitly. However, a powerful implicit compulsion comes from the combined force of two aspects: (1) before generating the EOS token (i.e., End-Of-Sequence Token), all the predicate constraints should be satisfied (you can imagine, in the training data at the position of EOS token in the ground-truth output, all elements in the state matrix must be set to "satisfied"); (2) The EOS token also should be generated within a limited decoding steps (any training data is of finite length). Such a soft way of combining symbolic operators (good at logical and mathematical calculations ) and neural operators (good at wording and phrasing) can retain their respective strengths to the utmost extent.

\section{Experiment}
We test the proposed method on two kinds of generation tasks: controllable text generation and general text generation. For controllable text generation, we choose ROCStories to verify that our method could generate a text under complex fine-grained control instruction.  We leverage commonsense generation and document-level machine translation to prove that NRETM can efficiently integrate prior knowledge into seq2seq models towards superior generation performance. 

\subsection{Controllable ROC Stories}
In order to fully tap the potential of our framework in executing complex control sentences (defined in the form of predicate logic), we choose an open-domain text generation benchmark ROCStories~\cite{mostafazadeh-etal-2016-corpus} which is a corpus of five-sentence stories that capture a rich set of causal and temporal commonsense relations between daily events, making them a good resource for training text generation models.

\textbf{Dataset} The ROCstories dataset consists of 98,162 five-line stories for training, and 1871 stories each for the development and test sets. Following~\cite{peng-etal-2018-towards}, we construct storyline, which is a (un)ordered sequence of phrases that are mentioned in the story, using RAKE algorithm~\cite{rose2010automatic}.

\textbf{Predicate Logic Formulation} 
As shown in table~\ref{rocexperiments}, We have experimented with various storyline constraints with increasing difficulties: (1) Generate a story in the order of the storyline $w_1$, $w_2$, $w_3$, $w_4$. (2) Generate a story with storyline $w_1$ in the $i^{th}$ sentence and $w_2$ in the $j^{th}$ sentence. (3) Generate a story with storyline $w_1$ in the $i^{th}$ sentence which has $l_i$ words and $w_2$ in the $j^{th}$ sentence whose length is $l_j$. (4) Generate a storyline $w_1$ in the $i^{th}$ sentence which has $l_i$ words or $s_i$ stop words and $w_2$ in the $j^{th}$ sentence that does not mention $w_3$ (5) Generate a storyline $w_1$ in the $i^{th}$ sentence which has $l_i$ words or $s_i$ stop words and $w_2$ in the $j^{th}$ sentence 
which has $l_j$ words or $s_j$ stop words. 

\textbf{Baselines and Metrics} We use the T5-Base model as baseline which takes the given predicate logic constraints as inputs and generates the five-sentence stories. We combine our proposed \mf module with the T5-Base model to track the execution of the constraints. We report \emph{Constraints Success Ratio} (CSR) and ROUGH-L (R-L) on the ROCStories Test Split to evaluate our \mf model.

\textbf{Main Results} As shown in Table~\ref{rocexperiments}, in all five predicate logic constraints, compared to the T5 model, the \mf model achieves higher Constraint Success Ratio and maintains a similar level of ROUGH-L, showing that the \mf model can be flexibly controlled without loss of generated text quality. The gap in CSR between the T5 and \mf model is moderate in the first two constraints with simple token permutations. However, the success ratio of T5 model drops significantly given constraints that requires long-range numerical tracking (e.g., sentence length and the count of stop words).

\begin{table}[!ht]
\centering
\caption{Controllable ROCStories Experiment Results.}
\setlength\tabcolsep{2pt}
\renewcommand{\arraystretch}{1.2}
\begin{tabularx}{\textwidth}{cYYY}
\toprule
Predicate Logic Constraint  & Model & CSR & R-L \\ \midrule
\multirow{2}{*}{$\mathrm{InSen}(w_1, y^i) \land \mathrm{InSen}(w_2, y^j)$}                                       & T5   &  98.7  &     33.1  \\
    & \mf &  98.8  &   33.3    \\ \midrule
\multirow{2}{*}{$\mathrm{Order}(w_1, w_2) \land \mathrm{Order}(w_2, w_3) \land \mathrm{Order}(w_3, w_4)$}   
& T5   &  97.0  &  49.6     \\  
& \mf &   97.4 &    49.5   \\ \midrule
\multirow{2}{*}{$\mathrm{InSen}(w_1, y^i) \land \mathrm{InSen}(w_2, y^j) \land \mathrm{Len}(y^j, l_j) \land \mathrm{Len}(y^i, l_i)$} &
  T5 & 18.5 & 32.9 \\ 
  & \mf &  84.5  &   32.6    \\ \midrule
\multirow{2}{*}{\begin{tabular}{@{}c@{}}$\mathrm{InSen}(w_1, y^i) \land \mathrm{InSen}(w_2, y^j) \land (\lnot \mathrm{InSen}(w_3, y^j))$ \\ $\land (\mathrm{Len}(y^i, l_i) \lor \mathrm{StopWordCount}(y^i, s_i))$\end{tabular}} & T5 & 23.5 & 33.0 \\ 
& \mf &  49.4  &33.1       \\ \midrule
\multirow{2}{*}{\begin{tabular}{@{}c@{}}$\mathrm{InSen}(w_1, y^i) \land \mathrm{InSen}(w_2, y^j) \land (\mathrm{Len}(y^i, l_i) \lor \mathrm{StopWordCount}(y^i, s_i))$ \\ $\land (\mathrm{Len}(y^j, l_j) \lor \mathrm{StopWordCount}(y^j, s_j))$\end{tabular}} & T5 & 11.0  & 32.9 \\ 
& \mf & 35.9   &    33.0   \\
\bottomrule
\end{tabularx}
\label{rocexperiments}
\end{table}

\subsection{Commonsense Generation}
C{\small OMMON}G{\small EN} is a generation benchmark dataset target explicitly test machines for the ability of generative commonsense reasoning. Given a set of common concepts the task is to generate a coherent sentence describing an everyday scenario using these concepts.

\textbf{Dataset} The C{\small OMMON}G{\small EN} dataset consists of 35,141 concept-sets (32,651 in \emph{train}, 993 in \emph{val}, 1,497 in \emph{test}) associated with 77,449 sentences. 

\textbf{Predicate Logic Formulation} The input is an unordered set of $n$ concepts $\mathbf{x} = \{x_i\}_{i=1}^n$. The expected output is a coherent, grammatical sentence $\mathbf{y}$ that describes a common scenario using all given concepts in $\mathbf{x}$. From the expectation of C{\small OMMON}G{\small EN}, one easily obtained prior knowledge is that each $x_i$ must appear in output $\mathbf{y}$. The corresponding predicate logic constraint $P_c$ is:
\[P_c = \land_{i=1}^n \big( \text{Copy}(x_i)\big)\]
where $\mathbf{y}$ will appear by default, for the sake of brevity, we have omitted $\mathbf{y}$ in predicate $\text{Copy}$. Another prior knowledge comes from the observation that generating $\mathbf{y}$ requires giving the correct morphological inflections of the concept word rather than copy it as it is. Let $\tilde{x}_i = \{\tilde{x}_k^i\}_{k=1}^{|\tilde{x}_i|}$ denote all inflections of $x_i$. $\mathbf{y}$ covers concept $x_i$, if at least one of $\{\tilde{x}_k^i\}_{k=1}^{|\tilde{x}_i|}$ appears. The constraint $\hat{P}_c$ is:
\[\hat{P}_c = \land_{i=1}^n \big( \lor_{j=1}^{|\tilde{x}^i|} \text{Copy}(\tilde{x}^i_j)\big)\]

\textbf{Baselines and Metrics} The standard pipeline of approaching this problem is to consider it as a conditional sentence generation task. We experiment with the recent pretrained language models, including T5-Base and T5-Large\cite{raffel2019exploring}. All models are finetuned with their default hyperparameters. We equip \mf into the T5-Large and T5-Base model to incorporate $P_c$ and $\hat{P}_c$ respectively (+ \mf $P_c$) (+ \mf $\hat{P}_c$). Grid Beam Search (GBS)~\cite{hokamp2017lexically} is a well-designed decoding method that ensures the generation model satisfies the lexical constraints. We only apply GBS to the T5-Base model due to the memory constraint (+ $\text{G}$). Following the suggestions in~\cite{lin-etal-2020-commongen}, we use CIDEr~\cite{vedantam2015cider} and SPICE~\cite{anderson2016spice} to automatically assess the quality of generated texts. We calculate constraint satisfaction for all constraints (ALL), novel constraints (Novel) and seen constraints (Seen).

\textbf{Main Results} Table~\ref{commongenresult} shows that the \mf model improves the constraint satisfaction over the baselines for all cases, achieving close to 100\% (i.e., 99.3\% and 99.4\%). While GBS achieves perfect constraint satisfaction (i.e., 100\%), doing so significantly degrades the output text quality (more than 50 CIDEr), indicating the necessity integrating piror knowledge in training rather than inference. In addition, both prior knowledge $P_c$ and $\hat{P}_c$ have a positive effect on our model, improving our T5-large baseline by 3.7 and 11.4 CIDEr score, respectively. Finally, our \emph{T5-Large + \mf $\hat{P}_c$} model outperforms the previous state-of-the-art result~\cite{KGBART}, which integrates the \emph{ConceptNet}~\cite{10.5555/3298023.3298212} into the BART model, suggesting that our incorporated task-specific prior knowledge could be as powerful as knowledge from large-scale hand-crafted corpus. We calculate the average growth of CIDEr score for the four improvement directions: (1) double the parameters (from \emph{T5-Base} to \emph{T5-Large}) (2) add more piror knowledge (from $P_c$ to $\hat{P}_c$) (3) add symbolic operator (only increase 0.9\% parameters by adding \mf). The growth from (1) to (3) is 4.06,  6.55, 4.3. Unexpectedly, adding \mf into T5-Base model is better than adding another T5-base, only adding a new piror knowledge in \mf significantly outperforms doubling the parameters. All of the above shows how potential it is to find a method that could execute multiple rules effectively.

\begin{table}[!t]
\centering
\vspace{-3mm}
\caption{Experiment Results on Commonsense.}
\footnotesize
\setlength\tabcolsep{2pt}
\renewcommand{\arraystretch}{1.2}
\begin{tabularx}{0.7\textwidth}{lYYYYY}
\bottomrule
&&&  \multicolumn{3}{c}{Constraint}\\
\multirow{-2}{*}{Method}  &  \multirow{-2}{*}{CIDEr} &  \multirow{-2}{*}{SPICE} & Seen& Novel  & ALL  \\ 
\midrule
T5-Base & 159.5 & 31.9 & 95.4 & 93.3 & 94.0 \\
T5-Base + G & 110.7 & 27.8 & 100 & 100 & 100 \\
T5-Base + \mf $P_c$ & 164.4 &  32.1 & 95.7 & 94.6 &  95.6 \\
T5-Base + \mf $\hat{P}_c$ & 169.8 & 32.7 & 99.5 & 99.1 & 99.3 \\
\midrule
T5-Large & 163.6 & 32.4 & 94.4 & 93.0 & 93.9 \\
T5-Large + \mf $P_c$ & 167.3 & 33.0 & 93.9 & 93.8 & 93.9 \\
T5-Large + \mf $\hat{P}_c$ & 175.0 & 33.8 & 99.6& 99.2 & 99.4  \\
\midrule
KGBART~\cite{KGBART} & 168.3 &  32.7 &  -  & - & 98.6 \\
\bottomrule
\end{tabularx}
\label{commongenresult}
\end{table}

\subsection{Document-Level Machine Translation}
Document-level machine translation tasks is a general text generation task, where the goal is to translate segments of text that contain more than one sentence (up to an entire document). 

\textbf{Dataset}  We conduct experiments on the common document-level MT dataset: TED15 Zh-En which is from IWSLT 2014 and 2015~\cite{cettolo2012wit3,cettolo2015iwslt}. Following~\cite{miculicich-etal-2018-document}, we use 2010-2013 TED as the test set. 

\textbf{Predicate Logic Formulation} The input is an ordered set of $n$ sentences in the source language that form a document $\mathbf{x}=\{x^i\}_{i=1}^n$, the expected output is a translated document $\mathbf{y}=\{y^i\}_{i=1}^n$ = in the target language. We observed that neural model is prone to sentence correspondence confusion (the $i^{th}$ sentence in source document is translated as the $j^{th}$ sentence in target document) when doing document-level translation. To alleviate this problem, we propose incorporating Doc-mBART25 with prior knowledge: each source sentence should be translated only once. It is formulated as:
\begin{equation}
\text{TranslatedOnce}(x^i)=\left\{
\begin{array}{rcl}
2 & & { \theta(\mathbf{y}_{:t})~>~i }  \\
1 & & {\theta(\mathbf{y}_{:t})~=~i}\\
0 & & {\theta(\mathbf{y}_{:t})~<~i} \end{array} \right.
\label{mfdefinedmt}
\end{equation}
where $\theta(\cdot)$ returns the line number of $y_t$ in $\mathbf{y}$, as t is monotonic during generation, the status only set to be 2 once.
The predicate logic constraint $P_c$ of this task can be formulated as:
\[P_c = \land_{i=1}^n \big( \text{TranslatedOnce}(x^i)\big)\]

\textbf{Baselines and Metrics} We combine our \mf $P_c$ component with the Doc-mBART25 model proposed in~\cite{lewis-etal-2020-bart} which is a state-of-the-art multilingual pre-trained language model. We compare this model with the state-of-the-art non-pretraining and pretraining approaches, including HAN (Hierarchical Attention Networks)~\cite{miculicich-etal-2018-document}, Doc-mBART25 and Sen-mBART25~\cite{lewis-etal-2020-bart}. When implementing our model, we use the same pre-processing method, blocks segmentation strategy and beam search setting as~\cite{lewis-etal-2020-bart}. Unlike \cite{lewis-etal-2020-bart}, We use both document-level (d-BLEU) and sentence-level (s-BLEU) to measure the similarities between generated target document and the source document. Besides, we also calculate the 
Sentence Aligned Ratio (SAR) between the source and target documents.

\begin{wraptable}{r}{0.54\textwidth}
\caption{Model Performance on TED15 Zh-En Test.}
\label{nmt}
\resizebox{0.54\textwidth}{!}{
\begin{tabular}{cccc}
\\\toprule
Model      & s-BLEU & d-BLEU & SAR \\ \midrule
Doc-mBART25 + \mf $P_c$ & 24.9   & 30.4   & 100 \%   \\ 
\midrule
Doc-mBART25~\cite{lewis-etal-2020-bart}      & 23.8      & 29.6   & 98.7 \%         \\ 
Sen-mBART25~\cite{lewis-etal-2020-bart} & - & 28.4 & - \\ 
HAN~\cite{miculicich-etal-2018-document}        & -      & 24.0   & -  \\ 
\bottomrule
\end{tabular}
}
\end{wraptable}

\textbf{Main Results} Table~\ref{nmt} shows that the \mf $P_c$ component helps the Doc-mBART25 model to better capture the sentence-level corresponding relationship between the source and target documents. In particular, sentence-level alignment ratio is improved from 98.7\% to 100\%. The improvement in s-BLEU (+ 1.1 BLEU) also confirms that our final Doc-mBART25 + \mf $P_c$ model learns to translate sentences based on the sentence order in source documents.

\subsection{Discussion}

\begin{wraptable}{r}{0.35\textwidth}
\centering
\vspace{-5mm}
\caption{Novel Sentence Index Experiment. MR for Mention Ratio.}
\label{zeroshot}
\resizebox{0.35\textwidth}{!}{
\begin{tabular}{cccc}
\\\toprule  
model   & CSA  & MR   & R-L  \\ \midrule
T5      & 19.7 & 95.7 & 30.0 \\ \midrule
\mf & 97.7 & 98.3 & 33.3 \\ 
\bottomrule
\end{tabular}
}
\end{wraptable} 

\textbf{Zero-Shot Execution} In Table~\ref{rocexperiments}, we show that simple sequence-to-sequence pre-trained language model (i.e., T5) cannot handle complicated and fine-grained constraints even after fine-tuning. Here, we further demonstrate that \mf model is capable to handle zero-shot rule execution. We train the T5 and \mf model to only mention storyline in the $3^{rd}$, $4^{th}$ and $5^{th}$ sentence and test these models to mention storyline in the first and second sentence of the whole story. As shown in Table~\ref{zeroshot}, although both T5 and \mf model mention most of the storyline (95.7\% and 98.3\% respectively) in the generated story, the T5 model only mention 19.7\% of storyline in the correct sentence and the \mf model makes 97.7\% of storyline correct. This is becuase the seuqence-to-sequence T5 model cannot recognize the novel sentence index ``1'' and ``2'' during the generation. The logic tracker helps the \mf model to generalize to handle these cases. 

\textbf{Case Study} More generation results of our models can be found in the Supplementary Material A.1.

\section{Related Work}
We are mainly related to two lines of research work in the field of text generation: constrained generation and prior knowledge integration. Some of the earliest works~\cite{ficler2017controlling,yu2017seqgan,kikuchi2016controlling} involve controllable generation methods where the generators are trained on text data with the labeled target attributes. CTRL\cite{keskar2019ctrl}, PPLM\cite{dathathri2019plug} and CoCon\cite{chan2020cocon} are recent approaches that built on the transformer-based large-scale pretrained LMs, they pay more attention on controlling high-level attributes,phrases and keywords. While NRETM focuses on controlling text generation to follow arbitrary logical constraints, leading to a fine-grained control. Recently, GDC\cite{khalifa2020distributional} permits to specify both pointwise and distributional constraints over the target LMs. In the field of constrained generation closest to our work is NEUROLOGIC\cite{lu2020neurologic} which seeks to generate fluent text while satisfying complex lexical constraints (in a predicate logic form). While NEUROLOGIC’s design also enables controlled generation with logical constraints, our approach leverages executable programs as logical operators and the dynamic tracking mechanism to realize a highly customizable and scalable predicate logic function system rather than simply controlling what words should and should not appear in the output. Existing efforts~\cite{zhang2018prior,baziotis2020language,zhang2017improving,tang2016neural,shen2015minimum} to incorporate prior knowledge into sequence-to-sequence framework either resort to modifying model architectures or designing training objectives. To the best of our knowledge, we first attempt to formalize the prior knowledge integration in seq2seq generation as text generation that conforms to predicate logic constraints.
\section{Conclusion and Future Work}
In this paper, we propose a unified controllable generation framework that leverages predicate logic constraints to implement efficient complex fine-grained control and scalable prior knowledge integration. We explore and compare two controllable strategies: dynamic tracking and static strategy, and show that the proposed dynamic tracking mechanism significantly outperforms the static ones. Empirical results on three benchmarks indicate that \mf could achieve accurate control and exhibits a superior generation ability over different tasks. Pre-trained models have been the dominant paradigm in natural language processing, and researchers resort to massive data and large-scale models to improve performance. We unify the rules used in various tasks into the form of predicate logic, provide the possibility to pretrain models on massive rules. In the future, we will explore pre-training large-scale neural rule-execution machine with massive rules and data.

\section*{Broader Impact}
Our work proposes a unified and scalable approach to efficiently perform fine-grained controllable text generation and incorporate multiple prior knowledge for superior text generation performance. This work uses story generation, machine translation, commonsense generation as applications to verify the effectiveness. However, it could be used in a broader range of more complex scenes (e.g., financial report generation) due to its advantages of combining multiple rules to generate high-quality texts. On the positive side, more efficient text generation can make these technologies more available to the general public. Our model verified on machine translation can benefit commercial and humanitarian translation services to help overcome language barriers. With these advantages also come potential downsides: powerful language models might be used to generate abuse, faked or misleading content in the news or on social media; they might pose safety concerns if they are used to generate harassing or hateful materials.  In order to mitigate these risks, AI systems could be leveraged to fight against misleading content and harassing material.

{\small
\bibliography{./nips2021.bib}
}


\clearpage
\section*{Appendix}

In this Appendix, we provide more details and examples for our proposed \mf models. A.1 provides the case study for three evaluation tasks. A.2 provides the definitions of six predicates used in this paper. A.3 provides some more analysis about predicate logic constraint and state flag as well as the actual state flag matrix used in the Commonsense Generation task (Sec 3.2) and Document-level Machine Translation task (Sec 3.3). A.4 provides more results regarding the controllable ROCStories Generation Experiments (Sec 3.1). Finally, A.5 provides implementation details of our models.

\subsection{Case Study}

In this case study, we select representative examples from our T5 baseline and \mf models. The \emph{Logic expression} is the actual predicate logic constraints that we use in the model. The \emph{Input expression} is the actual string that we feed into the T5 model simply because T5 vocabulary does not include $\land$, $\lor$, and $\neg$. 

\begin{table}[!ht]
\caption{Generated examples of the T5 and \mf model for the story generation task.}
\small
\centering
\begin{tabularx}{\linewidth}{rL}
\toprule
\multicolumn{2}{c}{{\bf Task 1: Story Generation}} \\
\midrule

Input expression: & $\mathrm{Order}$(hated, stupid) \&
$\mathrm{Order}$(stupid, insulting) \& $\mathrm{Order}$(insulting, punched) \\
Logic expression: & $\mathrm{Order}$(hated, stupid) $\land$
$\mathrm{Order}$(stupid, insulting) $\land$ $\mathrm{Order}$(insulting, punched) \\

\textbf{T5}: & I had a crush on a man. I told him I was stupid. \textcolor{red}{He hated me for insulting me. I punched him in the face.} He told me i was stupid. \\
\textbf{NRETM}: & I had a crush on a guy. He was very rude. \textcolor{red}{He hated me because i thought he was stupid and insulting me. I punched him in the face.} He slapped me. \\
\midrule
Input expression: & $\mathrm{InSen}$(walking around, 2) \& $\mathrm{InSen}$(school yard, 2) \\
Logic expression: & $\mathrm{InSen}$(walking around, $y^2$) $\land$ $\mathrm{InSen}$(school yard, $y^2$) \\ 

\textbf{T5}: & I went to the \textcolor{red}{park} today. \textcolor{red}{I was walking around the school yard}. I saw a squirrel in the yard. I saw a squirrel in the yard. I took a picture of the squirrel. \\
\textbf{NRETM}: & Sam was playing in the \textcolor{red}{yard}. \textcolor{red}{He was walking around in the school yard}. He saw a squirrel in the yard. He ran into the squirrel. Sam was able to retrieve the squirrel. \\
\midrule

Input expression: & $\mathrm{InSen}$(waitressing job, 3) \& ($\mathrm{Len}$(3, 16) || \textcolor{red}{$\mathrm{StopWordCount}$(3, 8)}) \& (not $\mathrm{InSen}$(tenacity, 3)) \& $\mathrm{InSen}$(talent agent, 4) \\
Logic expression: & $\mathrm{InSen}$(waitressing job, $y^3$) $\land$ ($\mathrm{Len}$($y^3$,16) $\lor$ \textcolor{red}{$\mathrm{StopWordCount}$($y^3$,8)}) $\land$ ( $\neg$ $\mathrm{InSen}$(tenacity, $y^3$)) $\land$ $\mathrm{InSen}$(talent agent, $y^4$) \\ 
\textbf{T5}: & Sarah had a crush on a girl named Sarah. Sarah s crush was Sarah's boyfriend. \textcolor{red}{Sarah's boyfriend \underline{had} \underline{a} waitressing job \underline{and} Sarah \underline{had} \underline{a} crush \underline{on} Sarah .} Sarah s boyfriend was a talent agent. Sarah's crush was sarah's boyfriend. \\
\textbf{NRETM}: & Sarah wanted to be a chef. but she didn't have the money to make a living. \textcolor{red}{She applied \underline{for} \underline{a} waitressing job \underline{but} she \underline{did} \underline{n't} get \underline{a} job \underline{at} \underline{all} .} Sarah was approached by a talent agent. Sarah was hired and now she is a chef. \\
\midrule
Input expression: & $\mathrm{InSen}$(longest grocery trip, 1) \& $\mathrm{Len}$(1, 12) \& $\mathrm{InSen}$(line forever, 4) \& $\mathrm{Len}$(4, 9) \\
Logic expression: & $\mathrm{InSen}$(longest grocery trip, $y^1$) $\land$ $\mathrm{Len}$($y^1$, 12) $\land$ $\mathrm{InSen}$(line forever, $y^4$) $\land$ $\mathrm{Len}$($y^4$, 9) \\ 
\textbf{T5}: & \textcolor{red}{Sam was on the longest grocery trip of his life.} He was so hungry he couldn't wait to get to the store. When he got to the store he had to wait in line forever. \textcolor{red}{Sam had to wait in line forever for his next grocery trip.} \\
\textbf{NRETM}: & \textcolor{red}{I was on the longest grocery trip I ever went on.} I had to wait in line for a long time. When I got there the line was long. \textcolor{red}{I had to wait in line forever again.} I was so glad I finally got to the store \\
\midrule
Input expression: & $\mathrm{InSen}$(could buy one, 2) \& ($\mathrm{Len}$(2, 13) || $\mathrm{StopWordCount}$(2, 6)) \& $\mathrm{InSen}$(would work, 4) \& $(\mathrm{Len}$(4, 10) || $\mathrm{StopWordCount}$(4, 7)) \\
Logic expression: & $\mathrm{InSen}$(could buy one, $y^2$) $\land$ ($\mathrm{Len}$($y^2$, 13) $\lor$ $\mathrm{StopWordCount}$($y^2$, 6)) $\land$ $\mathrm{InSen}$(would work, $y^4$) $\land$ ($\mathrm{Len}$($y^4$, 10) $\lor$ $\mathrm{StopWordCount}$($y^4$, 7)) \\ 
\textbf{T5}: & I needed a new computer. \textcolor{red}{I asked my mom \underline{if} I \underline{could} buy \underline{one}.} She said yes. \textcolor{red}{I asked her if it would work for me.} She said yes \\
\textbf{NRETM}: & I needed a new computer. \textcolor{red}{I went \underline{to} \underline{the} store \underline{to} see \underline{if} I \underline{could} buy \underline{one}.} I went to the store to buy a computer. \textcolor{red}{I found a computer that would work for me.} I bought the computer and was very happy. \\
\bottomrule
\end{tabularx}
\label{examplet1}
\end{table}

Table~\ref{examplet1} shows selected examples in Controllable ROCStories Generation task. This task is to show the controllability of our proposed \mf model. Sentences \textcolor{red}{in red} are the ones being controlled. In the first story, the T5 baseline model produces a short sentence and misses the order of storyline "stupid" which should appear after generating the storyline "hated". While our \mf model successfully completes all storylines in order. In the second story, the \mf model controls the story generation in a more coherent way than the T5 baseline model. Although both baseline and \mf model successful incorporate all given storylines, the T5 baseline model inconsistently generates ``school yard'' just after generating the ``park''. On the contrary, in the story generated by the \mf model,  Sam consistently stays in the ``yard''. In the third story, the length and stop word control force the \mf model to generate sentences with more details, while the T5 baseline simply repeats information from previous sentences. The \mf model successfully generates eight stop words in the third sentence, whereas the baseline model only generates six stop words (highlighted via \underline{underline}). In addition, the generated story from the \mf model has more rational plots than the one from the T5 model. In the fourth story, the length of the first and fourth sentences are controlled to be 12 and 9. The outputs of \mf model successfully obey these control constraints while the baseline model generates 11 and 13 tokens for the first and fourth sentences. In the last story, the second sentence generated by the \mf model successfully generates six stop words (highlighted via \underline{underline}). For this task, we are more concerned about the expression rate of predicate logic control constraints than the quality of the generated story. In addition to the case study, we have shown more quantitative analysis, and please refer to Sec. A.4 for details.

\begin{table}[!ht]
\caption{Generated Example of the T5 and \mf model in the Commonsense Generation task.}
\small
\centering
\begin{tabularx}{\linewidth}{rL}
\toprule
\multicolumn{2}{c}{{\bf Task 2: Commonsense Generation}} \\
\midrule
Input expression: & $\mathrm{Copy}$(stone) \&  \textcolor{red}{$\mathrm{Copy}$(explain)} \&  $\mathrm{Copy}$(knife) \&  $\mathrm{Copy}$(sharpen) \\
Logic expression: &  $\mathrm{Copy}$(stone) $\land$  \textcolor{red}{$\mathrm{Copy}$(explain)}  $\land \mathrm{Copy}$(knife) $\land \mathrm{Copy}$(sharpen) \\
\textbf{T5}: & a man is sharpening a knife on a stone \\
\textbf{NRETM}: & a man \textcolor{red}{explains} how to sharpen a knife on a stone \\
\midrule
Input expression: & $\mathrm{Copy}$(stand) \&  $\mathrm{Copy}$(map) \&  \textcolor{red}{$\mathrm{Copy}$(report)} \&  $\mathrm{Copy}$(front) \& $\mathrm{Copy}$(weather) \\
Logic expression: &  $\mathrm{Copy}$(stand) $\land$  $\mathrm{Copy}$(map) $\land$  \textcolor{red}{$\mathrm{Copy}$(report)} $\land$ $\mathrm{Copy}$(front) $\land$ $\mathrm{Copy}$(weather)  \\
\textbf{T5}: & map showing where the weather is standing at the front \\
\textbf{NRETM}: & a man stands in front of a map \textcolor{red}{reporting} the weather \\
\midrule
Input expression: & $\mathrm{Copy}$(put) \&  $\mathrm{Copy}$(lipstick) \&  $\mathrm{Copy}$(talk) \&  \textcolor{red}{$\mathrm{Copy}$(lip)} \\
Logic expression: &  $\mathrm{Copy}$(put) $\land$  $\mathrm{Copy}$(lipstick) $\land$  $\mathrm{Copy}$(talk) $\land$ \textcolor{red}{$\mathrm{Copy}$(lip)} \\
\textbf{T5}: & a woman puts lipstick on and talks about it \\
\textbf{NRETM}: & a woman is talking and putting lipstick on her \textcolor{red}{lips} \\
\midrule
Input expression: & $\mathrm{Copy}$(iron) \&  $\mathrm{Copy}$(straighten) \&  \textcolor{red}{$\mathrm{Copy}$(demonstrate)} \&  $\mathrm{Copy}$(hair) \\
Logic expression: &  $\mathrm{Copy}$(iron) $\land$  $\mathrm{Copy}$(straighten) $\land$  \textcolor{red}{$\mathrm{Copy}$(demonstrate)} $\land$ $\mathrm{Copy}$(hair) \\
\textbf{T5}: & a woman straightens her hair with an iron and \underline{shows} how to do it \\
\textbf{NRETM}: & a woman is \textcolor{red}{demonstrating} how to straighten her hair with an iron \\
\midrule
Input expression: & $\mathrm{Copy}$(bride) \&  $\mathrm{Copy}$(stand) \&  $\mathrm{Copy}$(bridesmaid) \&  $\mathrm{Copy}$(groomsman) \&  \textcolor{red}{$\mathrm{Copy}$(groom)} \\
Logic expression: &  $\mathrm{Copy}$(bride) $\land$  $\mathrm{Copy}$(stand) $\land$  $\mathrm{Copy}$(bridesmaid) $\land$ $\mathrm{Copy}$(groomsman) $\land$  \textcolor{red}{$\mathrm{Copy}$(groom)}\\
\textbf{T5}: & bride standing with her bridesmaids and groomsmen \\
\textbf{NRETM}: & the bridesmaids and groomsmen stand in front of the bride and \textcolor{red}{groom} \\
\midrule
Input expression: & $\mathrm{Copy}$(kitchen) \&  $\mathrm{Copy}$(watermelon) \&  $\mathrm{Copy}$(knife) \&  $\mathrm{Copy}$(cut)  \\
Logic expression: &  $\mathrm{Copy}$(kitchen) $\land$  $\mathrm{Copy}$(watermelon) $\land$  $\mathrm{Copy}$(knife) $\land$ $\mathrm{Copy}$(cut) \\
\textbf{T5}: & \textcolor{blue}{a knife cutting} a watermelon in a kitchen \\
\textbf{NRETM}: & \textcolor{blue}{a man cutting} a watermelon with a knife in the kitchen \\
\bottomrule
\end{tabularx}
\label{examplet2}
\end{table}

Table~\ref{examplet2} shows selected examples from our T5 baseline and \mf models in the Commonsense Generation task. Concepts that are missed in the baseline model outputs are \textcolor{red}{in red}. Words \textcolor{blue}{in blue} are the key difference between the output of baseline and \mf model. Note that we omit the synonyms for simplicity. Full Examples for this task can be found in Sec. A.3. Although the baseline model can correctly complete many 
$\mathrm{Copy}$ operations, it fails when the input combination is not commonly seen. For example, ``explain'' and ``knife'' in the first example. The baseline model also generates meaningless sentence when the inputs are complicated concepts combination in the second example. In addition, the baseline model cannot handle the case where some input concepts share the same prefix, such as ``groom'' and ``groomsman'' in the forth example. The baseline model seems to merge these morphological similar input concepts into a single concept and only mentions one of them in the outputs. Whereas the \mf model successfully completes all of $\mathrm{Copy}$ operations.

Table~\ref{examplet3} shows selected examples from our T5 baseline and \mf models in document-level machine translation.
In the first case, the mT5 baseline model produces duplicated sentences (``what happens when co2 emissions go up'', \textcolor{red}{in red}). As a consequence, it fails to translate a few important chunks in the source sentences (see \textcolor{blue}{in Blue}). This may due to the fact that the mT5 baseline model cannot handle long input documents well. While our \mf model translates all source sentences into fluent English. Sentences \textcolor{green}{in Green} are missed by the baseline model but successfully translated by the \mf model with the help of the predicate $\mathrm{translateOnce}$. In the second case, the baseline model skips the important word ``exchange'' (see \underline{underline} in the Input expression) in its translated text (highlighted \textcolor{red}{in red}). The \mf model accurately translates this sentence (highlighted \textcolor{blue}{in blue}). This shows that the \mf model is more focused on the current sentence than the T5 baseline model.
\begin{CJK*}{UTF8}{gkai}
\begin{table}[ht]
\caption{Generated Example of the mT5 and \mf model for document-level machine translation.}
\small
\centering
\begin{tabularx}{\linewidth}{rL}
\toprule
\multicolumn{2}{c}{{\bf Task 3: Document-level Machine Translation}} \\
\midrule
Input expression: & \textcolor{red}{当前二氧化碳巨大的排放量}\textcolor{blue}{将导致温度上升，温度的升高将引起一系列非常严重的后果，比如对天气的直接影响，或对生态系统的间接影响，生态系统无法应对剧烈变化的结果就是}生态系统的全面崩溃。二氧化碳排放增加和温度升高究竟成怎样的关系，两者间的正反馈效应为何？这中间有一些不确定因素，但不多。至于全球变暖的具体负面影响有多严重，这无法完全确定，但肯定极其严重。$\mathrm{translateOnce}$(1) \& $\mathrm{translateOnce}$(2) \& $\mathrm{translateOnce}$(3) \\
\textbf{mT5}: & \textcolor{red}{what happens when co2 emissions go up -- what happens when co2 emissions go up --} is that the system collapses. what's the positive feedback between co2 emissions and temperature rising, and what's the negative feedback between co2 emissions and temperature rising?there's some uncertainty about how severe it is, but it's certainly severe. \\
\textbf{NRETM}: & what's going to happen when co2 emissions go up \textcolor{green}{will cause temperatures to rise with a range of very severe consequences, such that, either directly or indirectly}, ecosystems will collapse. there's some uncertainty about how this relationship, the positive feedback, between co2 emissions and temperature rise will work. how severe it will be, it's not entirely clear. \\
\midrule
Input expression: & 我是跟着50个去往贾拉拉巴德战役的反叛者乘着卡车一起来的，当时我19岁，是一位住在佛罗里达州杰克逊维尔的素食主义者兼冲浪爱好者。\textcolor{red}{我用我的黑色匡威的低邦鞋 \underline{换了}一双棕色的皮拖鞋，并且对着我看不太清的政府的坦克发了一枚火箭。}这是我第一次到阿富汗。$\mathrm{translateOnce}$(1) \& $\mathrm{translateOnce}$(2) \& $\mathrm{translateOnce}$(3) \\
\textbf{mT5}: & I was riding a truck with 50 rebels who had gone to Jalabard to fight the Taliban, and I was 19 years old, a vegetarian and surfer in Jacksonville, Florida. \textcolor{red}{I took my black shoe boots and a brown leather shoe, and I threw a rocket at a government tank that I couldn't see.} This was my first trip to Afghanistan. \\
\textbf{NRETM}: & I was accompanied by 50 counter-insurgents on a truck heading to Jamalabad, Florida, when I was 19 years old, a vegetarian and surfers living in Jacksonville, Florida. \textcolor{blue}{I exchange brown leather sandals with my black jackets, and I fired a rocket at the tanks of the government that I didn't see.} This was my first time in Afghanistan \\
\bottomrule
\end{tabularx}
\label{examplet3}
\end{table}
\end{CJK*}

\clearpage

\subsection{Definition of Predicates}

\begin{figure}[!ht]
\centering
\includegraphics[width=0.75\textwidth]{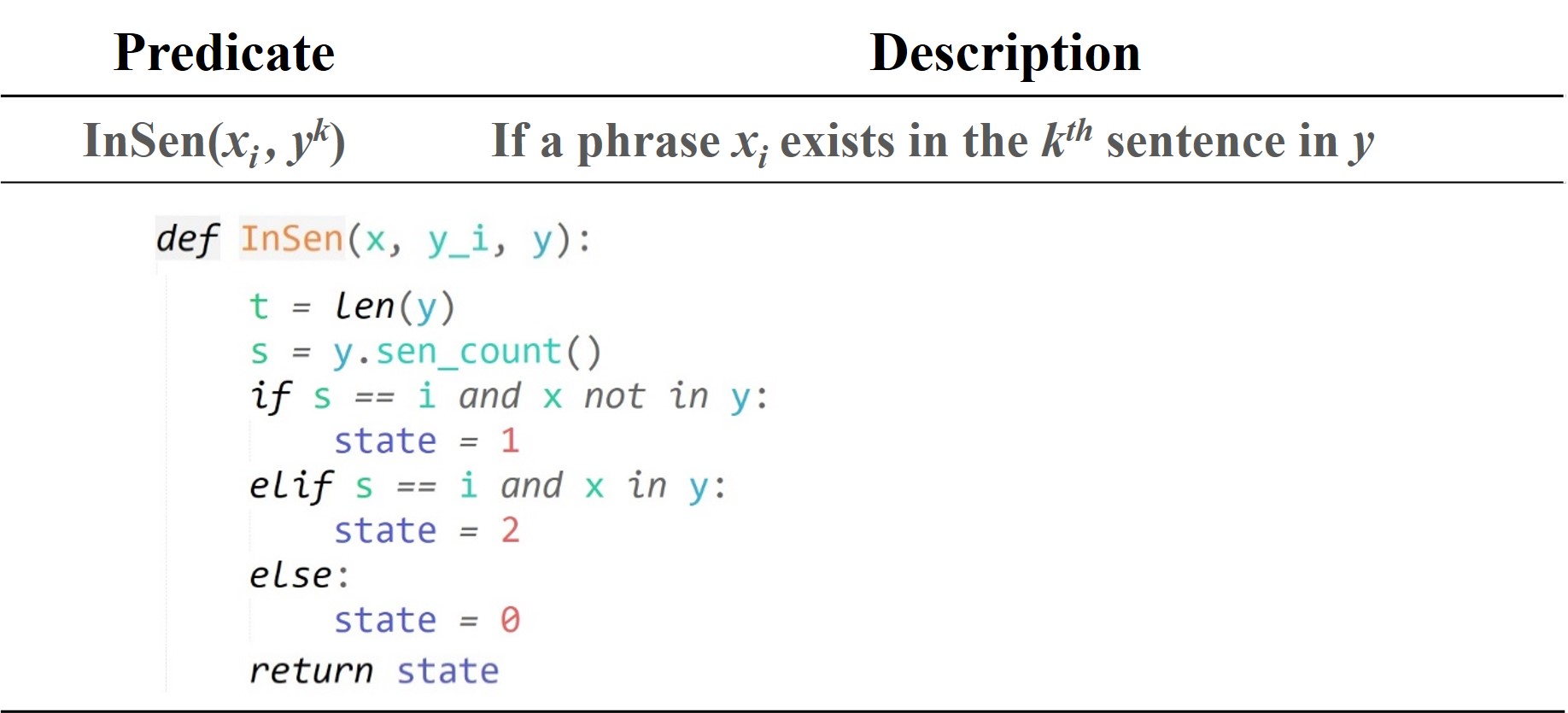}
\caption{The definition of predicate $\mathrm{InSen}$. The State 1 starts when the text generators start to generate $k^{th}$ sentence. This informs the model that it is possible to mention $x_i$ in the outputs.}
\label{example1}
\end{figure}

\begin{figure}[!ht]
\centering
\includegraphics[width=0.75\textwidth]{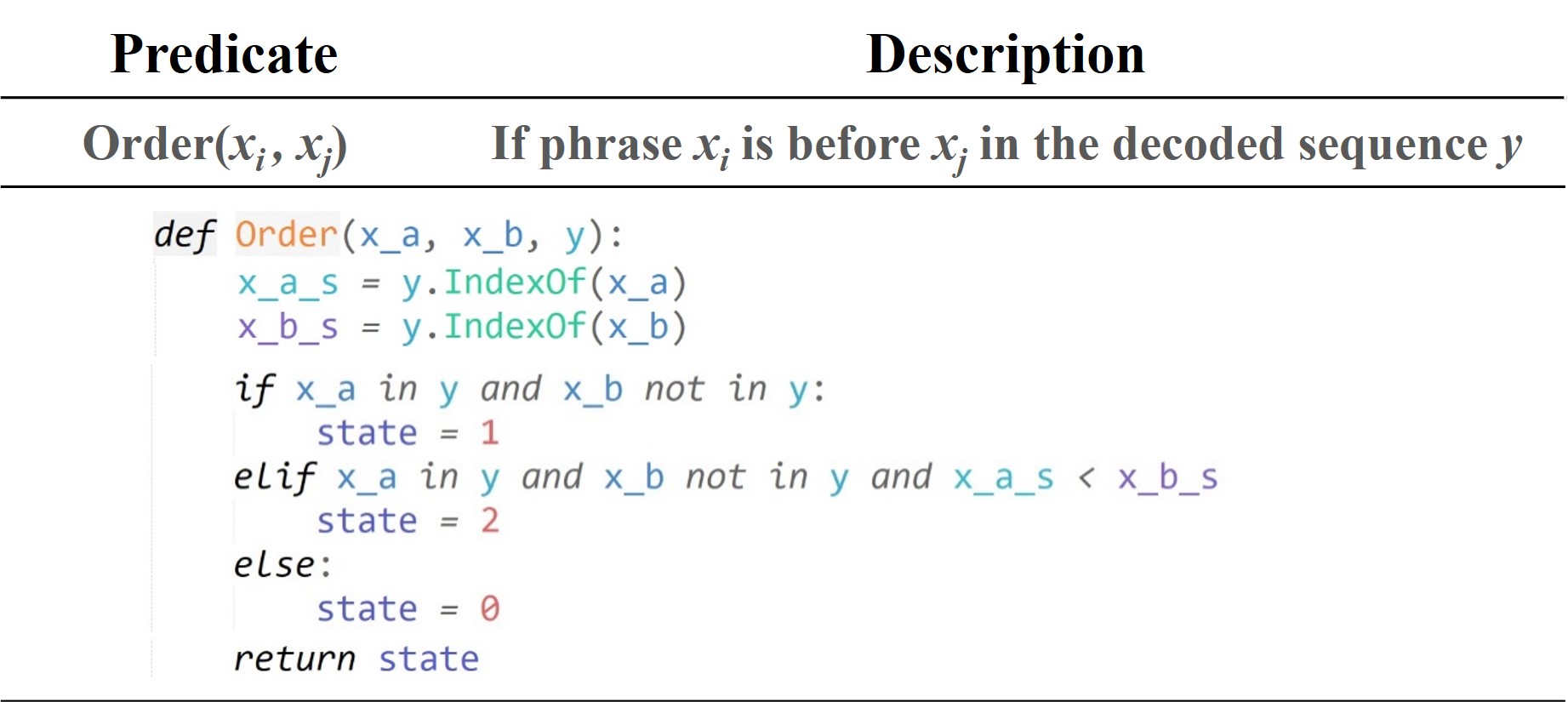}
\caption{The definition of predicate $\mathrm{Order}$. The State 1 starts when the previous element $x\_a$ has already been mentioned in the outputs. This informs the model to mention $x\_b$ next.}
\label{example2}
\end{figure}

\begin{figure}[!ht]
\centering
\includegraphics[width=0.75\textwidth]{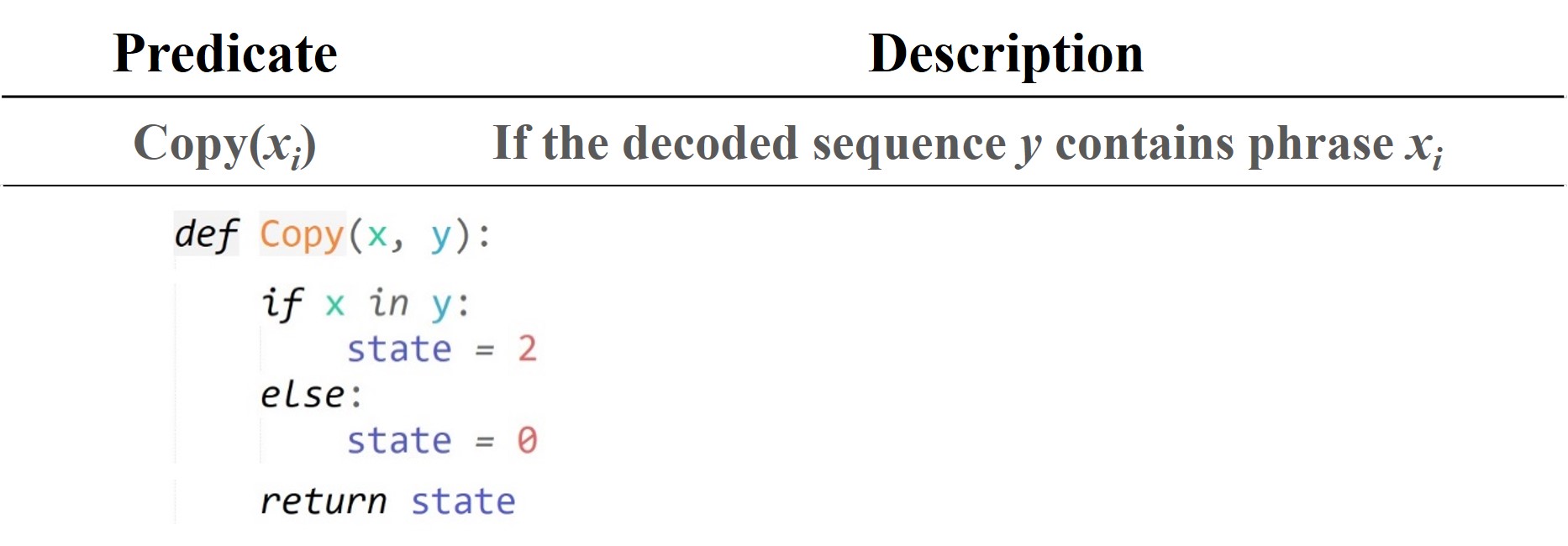}
\caption{The definition of predicate $\mathrm{Copy}$. There is no State 1 in the definition of $\mathrm{Copy}$ because there is no ``partial copy'' status.}
\label{example3}
\end{figure}

\begin{figure}[!ht]
\centering
\includegraphics[width=0.75\textwidth]{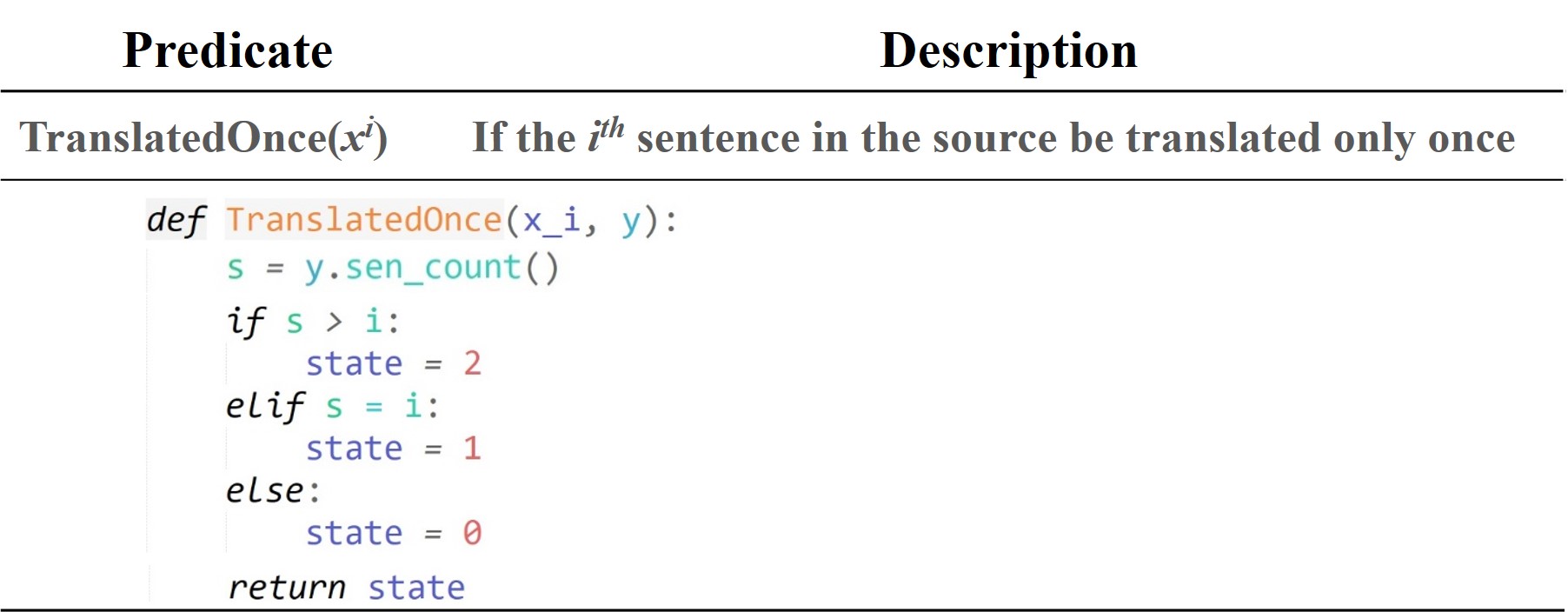}
\caption{The definition of predicate $\mathrm{TranslatedOnce}$. The State 1 starts when $i^{th}$ sentence is being translated. This informs the model should pay attention to which source sentence.}
\label{example4}
\end{figure}

\begin{figure}[!ht]
\centering
\includegraphics[width=0.75\textwidth]{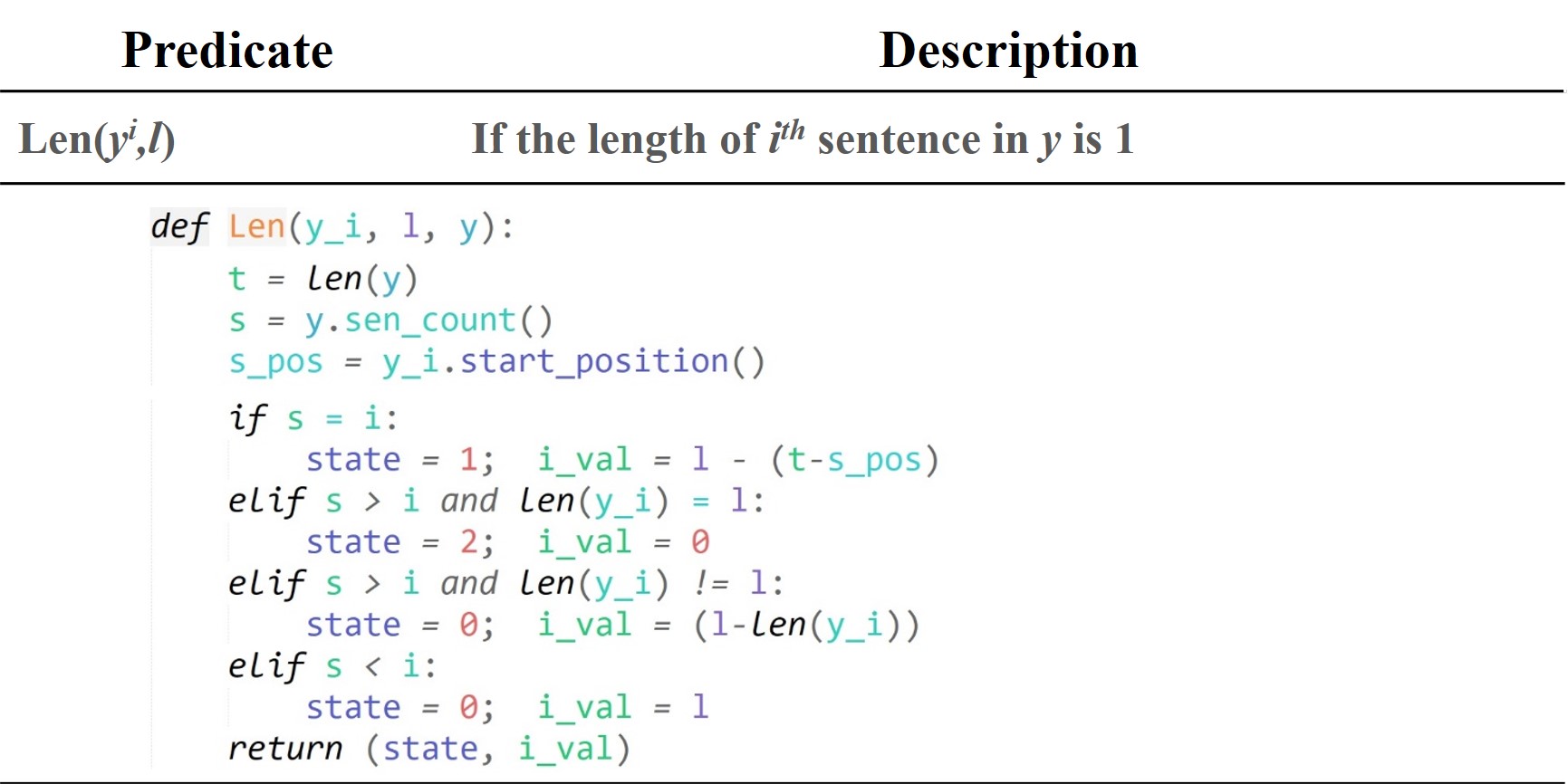}
\caption{The definition of predicate $\mathrm{Len}$. The State 1 starts when the text generator starts to generate $i^{th}$ sentence. We also explicitly inform the model of how many tokens are remaining for the current sentence. So they have State 1 with additional information $i\_val$. Figure~\ref{minimf} shows the actual state matrix of this predicate.}
\label{example5}
\end{figure}

\begin{figure}[!ht]
\centering
\includegraphics[width=0.75\textwidth]{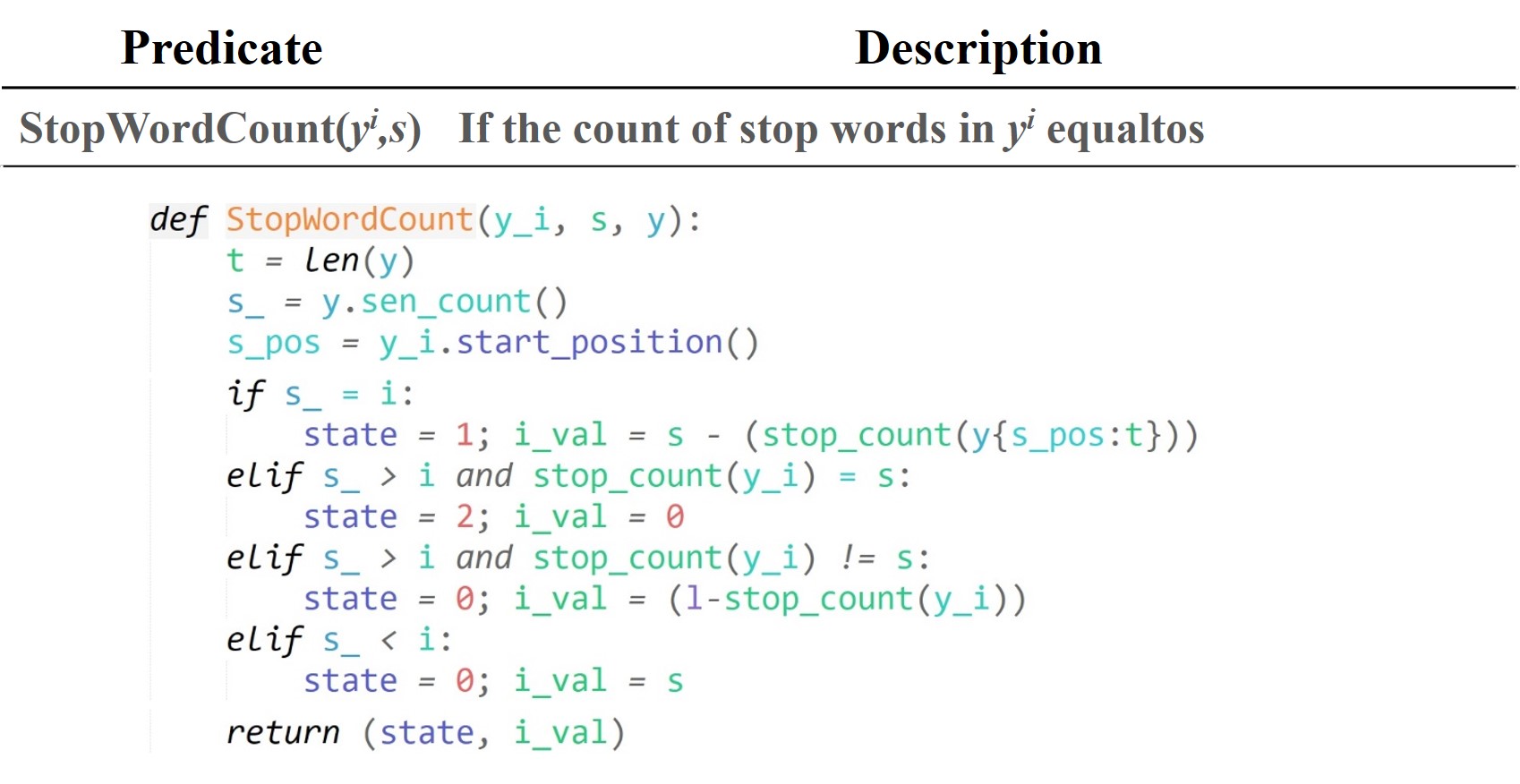}
\caption{The definition of predicate $\mathrm{StopWordCount}$. The State 1 starts when the text generator starts to generate $i^{th}$ sentence. We also explicitly inform the model of how many stop words are remaining for the current sentence. So they have State 1 with additional information $i\_val$.}
\label{example6}
\end{figure}

Figure~\ref{example1} to~\ref{example6} show the six kinds of predicates, including $\mathrm{InSen}$, $\mathrm{Order}$, $\mathrm{Copy}$, $\mathrm{TranslatedOnce}$, $\mathrm{Len}$ and $\mathrm{StopWordCount}$, used in our framework. The code in figures is the pseudo code of the logical operator (i.e., executable programs), in which y refers to $\mathbf{y_{:t}}$, y\_i is $y^i$, state is the default state status $\hat{q}_{it}^k$  and i\_val is the additional intermediate values $\bar{q}_{it}^k$ . In our experiments, all of these logical operators are implemented using the Python programming language, and their source codes are not directly visible to the neural text generators. They only communicate with the neural text generators using the state flags. All predicates have State 0, indicating unfinished status and State 2, indicating finished status. As discussed in Sec 2.4, State 1 is an optional predicate-specific state. We will introduce the definition and role of State 1 for each of the above predicate if it exists in the captions.

\clearpage

\subsection{Predicate Logic Constraint and State Flag}

As the predicate logic constraints $\mathcal{M}$ used in our framework could support arbitrary formats (i.e., predicates can be in any combination with logical words), a key challenge is mapping each state flag in the state matrix to the corresponding predicate in the logic expression. To tackle this challenge, we treat the predicate logic constraints (as "Logic expression" in tables of Sec. A.1) as the extra input (as "Input expression" in tables of Sec. A.1) to the encoder of sequence-to-sequence (\emph{S2S}) transformer-based text generators. In addition, we encode the state flags using a shallow transformer-based encoder with the same architecture. With the help of both positional embeddings in two modules, \mf could align the state flags in state matrix with predicates in logic expression to achieve successful control. For the state flag $q_{it}$, it keep tracks of the \emph{dynamic} progress of all predicates during text generation. Therefore, we only put the current progress of each predicate in $q_{it}$ and encode it using a shallow rule transformer encoder. For $\mathcal{M}$ with $n$ predicates $\mathcal{D}=\{U_i\}_{i=1}^n$, $q_{it}$ is the concatenation of $\{q_{it}^k\}_{k=1}^n$, where $q_{it}^k$ is the concatenation of default state status $\hat{q}_{it}^k$ and optional intermediate values $\bar{q}_{it}^k$. $\hat{q}_{it}^k$ is formulated as:
\begin{equation}
\hat{q}_{it}^k=\left\{
\begin{array}{rcl}
\text{N} & & {\text{variables of}~U_k~\text{do not contain}~x_i}  \\
0 & & {U_k~\text{is not satisfied}}  \\
1 & & {U_k~\text{is in progress}}\\
2 & & {U_k~\text{is satisfied}} \end{array} \right.
\label{mfdefinesupp}
\end{equation}

To show the importance of our proposed rule-execution tracking module, our baseline models also have access to the predicate logic constraints in their encoders in all of our evaluation tasks. The baseline model and our \mf model have the same amount of input information and only differ in whether equipped with the above rule-execution tracking module.

\begin{figure}[!ht]
\centering
\includegraphics[width=0.8\textwidth]{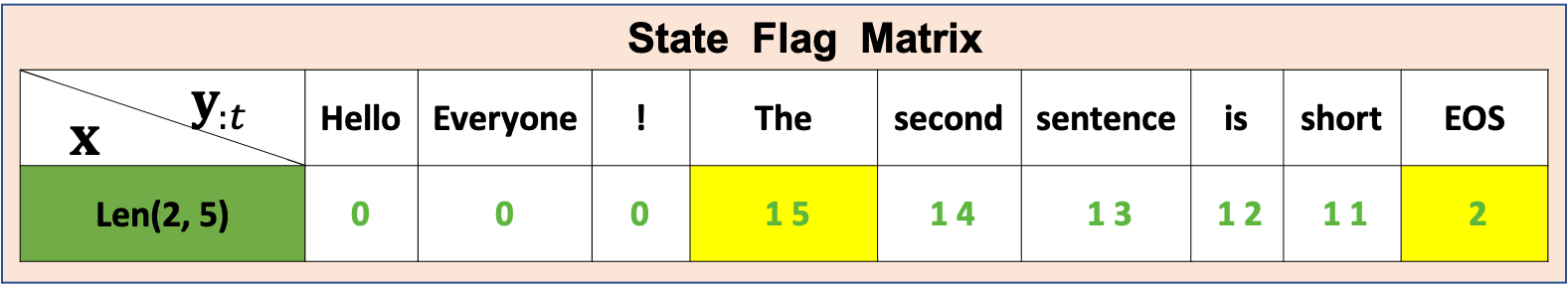}
\caption{The State Matrix for controlling the length of the second sentence. The yellow cell indicates the status update.}
\label{minimf}
\end{figure}

In Table 1, Sec 3.1, we control the length of arbitrary sentence for ROCStories generation. Figure~\ref{minimf} shows a minimal example of controlling the length of the second sentence. $\hat{q}_{it}^k$ is in status 0 when the model is generating the first sentence. As the model finishes the first sentence (i.e., after generating ``!''), $q_{it}^k$ is updated to ``1 5'' (see the definition of predicate $\mathrm{Len}$ in Figure~\ref{example5}). The $\hat{q}_{it}^k$ is finally updated to status 2 when finishing this sentence.

\begin{figure}[!ht]
\centering
\includegraphics[width=0.8\textwidth]{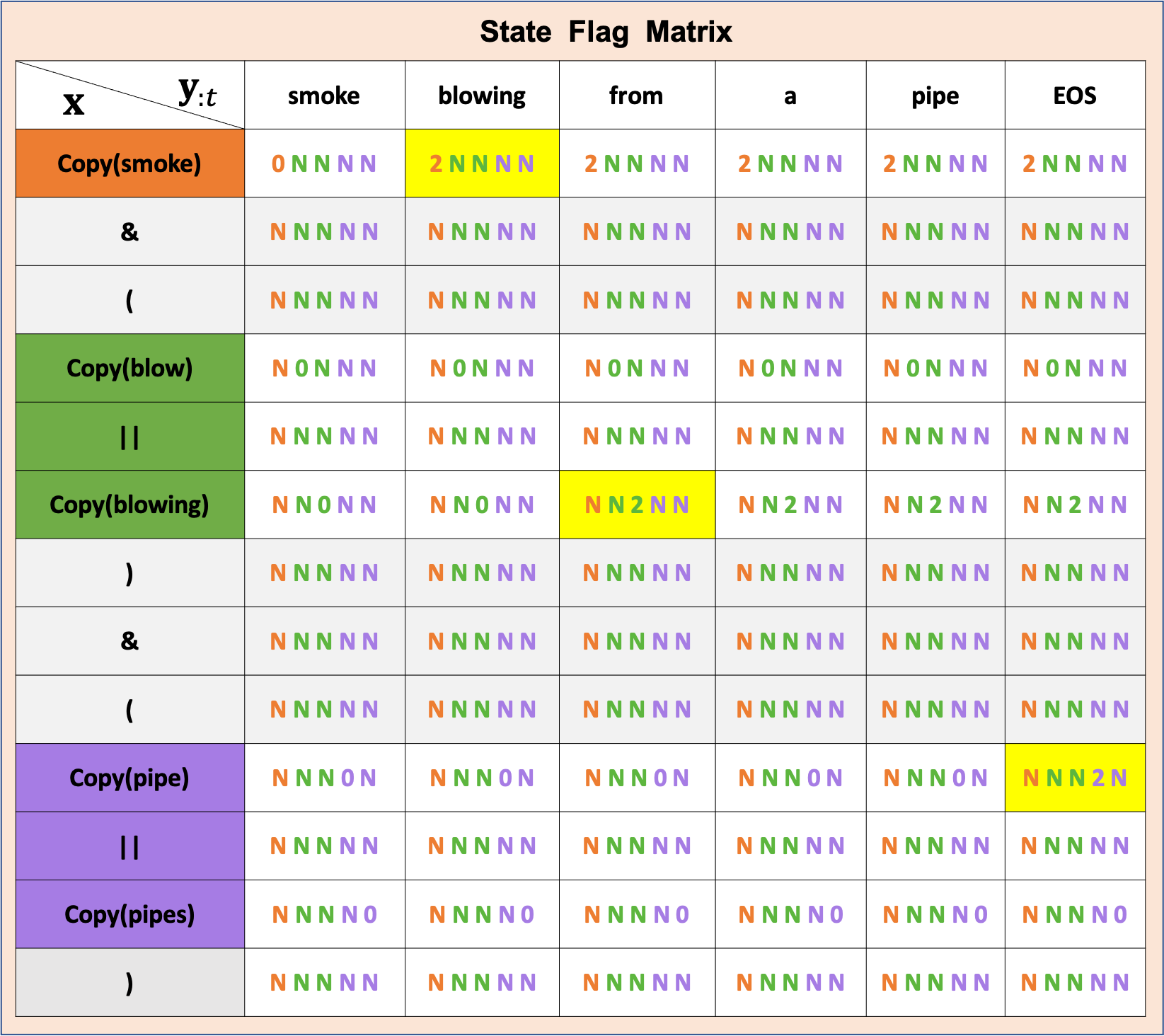}
\caption{The State Matrix for the Commonsense Generation task. In this task, the status is updated once the keywords or phrases are fully mentioned in the outputs.}
\label{commonsenmf}
\end{figure}

\begin{figure}[!ht]
\centering
\includegraphics[width=0.8\textwidth]{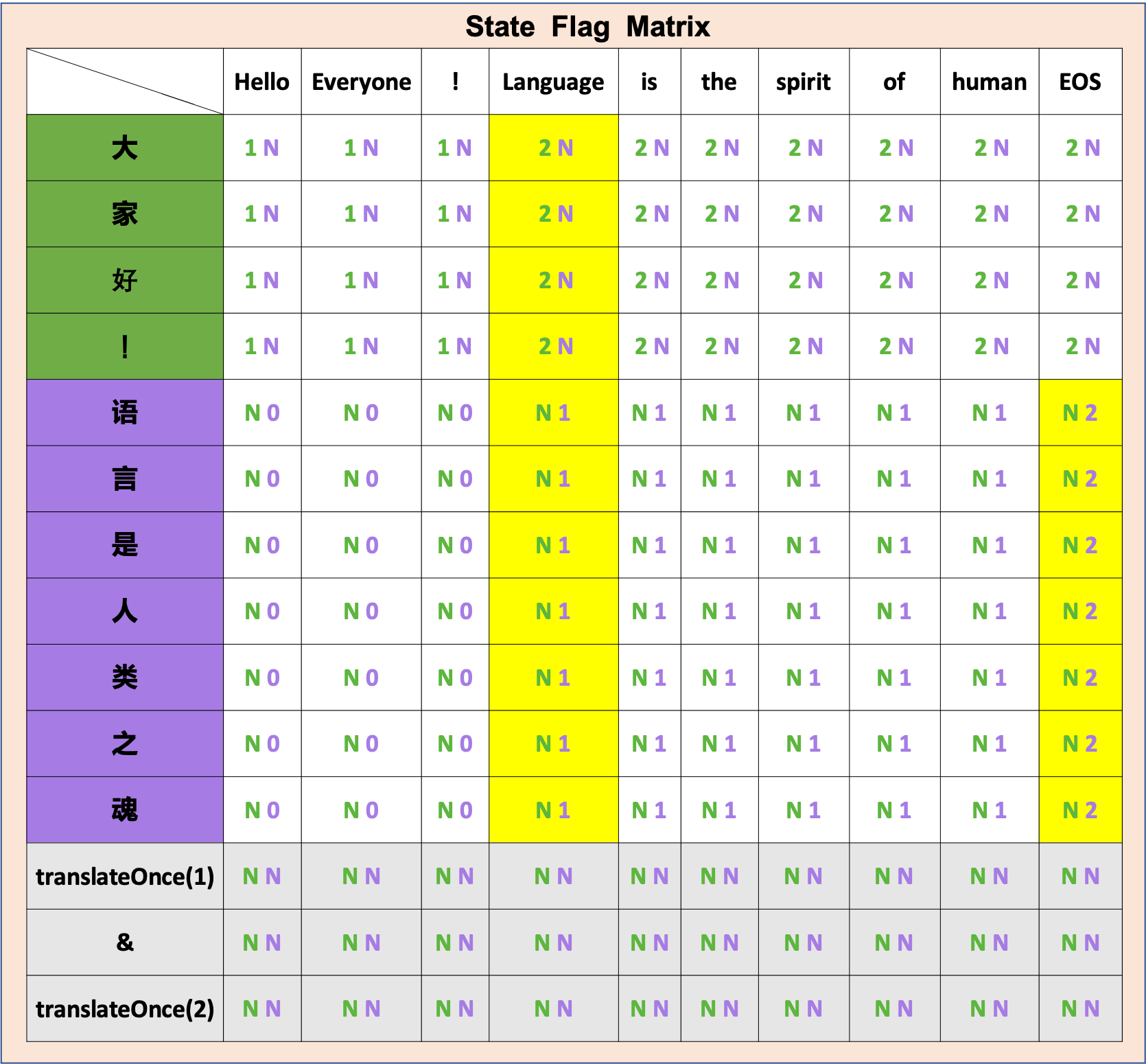}
\caption{The State Matrix for the Document-level Machine Translation. The yellow cell indicates the status update. In this task, the status is updated once one sentence is finished.}
\label{translationmf}
\end{figure}

We further show the actual state matrix used in the Commonsense Generation and the document-level machine translation task in Figure~\ref{commonsenmf} and~\ref{translationmf}, respectively. In both tasks, the number of $q_{it}^k$ is linear to the number of input concept words or the number of input source sentences. In our implementation, we compress $q_{it}^{k_1}$ and $q_{it}^{k_2}$ if they satisfy the following two conditions:
\begin{itemize}
  \item $U_{k_1}$ and $U_{k_2}$ are the same predicate.
  \item the variables of $U_{k_1}$ and $U_{k_2}$ are disjoint or the ``in progress'' period of $U_{k_1}$ and $U_{k_2}$ are not overlapped.
\end{itemize}
After the above compression, the length of $\{q_{it}^k\}$ in the Commonsense Generation task and Document-level Machine Translation task becomes one.

\clearpage

\subsection{More Results for Controllable ROCStories Generation Experiment}

In this section, we conduct more analysis in Table~\ref{rocexperimentsmore} and~\ref{rocexperimentsclause} for the rules with relatively low Constraint Success Ratio (CSR) in Table 1, Sec. 3.1 (Controllable ROCStories Generation Experiment). Specifically, they are:
\begin{itemize}
  \item $\mathrm{R}_1 = \mathrm{InSen}(w_1, y^i) \land \mathrm{InSen}(w_2, y^j) \land \mathrm{Len}(y^j, l_j) \land \mathrm{Len}(y^i, l_i)$;
  \item $\mathrm{R}_2 =  (\lnot \mathrm{InSen}(w_3, y^j)) \land (\mathrm{Len}(y^i, l_i) \lor \mathrm{StopWordCount}(y^i, s_i)) \land \mathrm{InSen}(w_1, y^i) \land \mathrm{InSen}(w_2, y^j)$;
  \item $\mathrm{R}_3 = (\mathrm{Len}(y^i, l_i) \lor \mathrm{StopWordCount}(y^i, s_i)) \land (\mathrm{Len}(y^j, l_j) \lor \mathrm{StopWordCount}(y^j, s_j)) \land \mathrm{InSen}(w_1, y^i) \land \mathrm{InSen}(w_2, y^j)$
\end{itemize}

The full CSR for $\mathrm{R}_1$, $\mathrm{R}_2$ and $\mathrm{R}_3$ is relatively low in Table 1, Sec 3.1. As all of them involve length and stop word control, we are interested in how accurate they are when they are allowed to make small control errors. We calculate Constraint Success Ratio with errors $\pm 1$ and $\pm 2$ in Table~\ref{rocexperimentsmore}. Under errors $\pm 1$, the CSR of \mf model is significantly improved. In $\mathrm{R}_3$, the CSR of the \mf model is improved from 35.9\% to 64.4\%. In three rules, the \mf model is 20\% - 30\% higher than the T5 baseline model. Under errors $\pm 2$, among all three rules, the lowest CSR for the \mf model is 82.1\%. The \mf model can still improve the CSR of baseline model by 6\% to 11\%. Note that $\pm 2$ is relatively large error gap in ROCStories dataset becuase each sentence only has, on average, 7.3 stop words. This explains the smaller CSR gap between the \mf and T5 baseline model in the CSR $\pm 2$ setup. In summary, the \mf model reasonably completes this controllable text generation task.  

\begin{table}[!ht]
\centering
\caption{Constraint Success Ratio with total length and stop word count errors $\pm 1$ and $\pm 2$.}
\setlength\tabcolsep{2pt}
\renewcommand{\arraystretch}{1.2}
\begin{tabularx}{0.7\textwidth}{cYYYY}
\toprule
\#R  & M &  CSR ($\pm 0$) & CSR ($\pm 1$) & CSR ($\pm 2$) \\ \midrule
\multirow{2}{*}{$\mathrm{R}_1$} 
& T5 & 18.5  & 67.6 & 89.2 \\ 
& \mf &  84.5    & 97.0 & 98.9  \\ \midrule
\multirow{2}{*}{$\mathrm{R}_2$} & T5 & 23.5 & 56.5 & 78.6 \\ 
& \mf &  49.4  &   76.1 & 89.9    \\ \midrule
\multirow{2}{*}{$\mathrm{R}_3$} & T5 & 11.0  & 44.9 & 76.2 \\ 
& \mf & 35.9   & 64.4 & 82.1 \\
\bottomrule
\end{tabularx}
\label{rocexperimentsmore}
\end{table}

We further break the CSR into predicate level in Table~\ref{rocexperimentsclause}. The \mf model achieves consistently higher CSR in all predicates than the T5 baseline model. Specifically, both T5 baseline and \mf model achieve near-perfect performance in the predicate $\mathrm{InSen}$. We believe that this is due to effect of the large-scale pre-training in the T5 model. The length control $\mathrm{Len}$ is more challenging than $\mathrm{InSen}$.
The \mf model achieves around CSR 90\%, while the baseline model only achieves CSR 38\%. This shows that simply feeding the target length to the \emph{S2S} encoder cannot properly control the output text length. When using the logical word $\lor$ in $\mathrm{R}_2$ and $\mathrm{R}_3$, the CSR for $\mathrm{Len}$ is around 50 - 60\%. Unlike $\mathrm{R}_1$ where the model is always trained to satisfy predicate $\mathrm{Len}$, in $\mathrm{R}_2$ and $\mathrm{R}_3$, only 67\% of the training data satisfy predicate $\mathrm{Len}$. The predicate $\mathrm{StopWordCount}$ is even more challenging: the \mf model only achieves 43.2\% and 28.6\% CSR in  $\mathrm{R}_2$ and $\mathrm{R}_3$, respectively. This may because the models have to distinguish between stop word tokens and non-stop word tokens.

\begin{table}[!ht]
\centering
\caption{Predicate-Level Constraint Success Ratio. $\mathrm{InS}$: $\mathrm{InSen}$; $\mathrm{L}$: $\mathrm{Len}$; $\mathrm{SWC}$: $\mathrm{StopWordCount}$;}
\setlength\tabcolsep{2pt}
\renewcommand{\arraystretch}{1.2}
\begin{tabularx}{\textwidth}{cYYYYYYY}
\toprule
\multicolumn{2}{c}{Model} &  \multicolumn{6}{c}{Predicate-Level CSR} \\ \midrule
\multirow{3}{*}{$\mathrm{R}_1$}
& Predicate & $\mathrm{InS}(w_1, y^i)$ & $\mathrm{InS}(w_2, y^j)$ & $\mathrm{L}(y^i, l_i)$ & $\mathrm{L}(y^j, l_j)$ & - & - \\ 
& T5 & 99.5  & 99.0 & 38.0 & 36.2 & - & - \\ 
& \mf &  99.5 & 99.3 & 89.7 & 91.2 & - & - \\ \midrule

\multirow{2}{*}{$\mathrm{R}_2$} 
& Predicate & $\mathrm{InS}(w_1, y^i)$ & $\mathrm{InS}(w_2, y^j)$ & $\lnot \mathrm{InS}(w_3, y^j)$ & $\mathrm{L}(y^i, l_i)$ & $\mathrm{SWC}(y^i, s_i)$ & - \\
& T5 & 99.6 & 99.1 & 99.1 & 23.5 & 18.8 & - \\ 
& \mf &  100  &  99.9 & 99.9 & 50.7 & 43.2 & -    \\ \midrule

\multirow{2}{*}{$\mathrm{R}_3$} 
& Predicate & $\mathrm{InS}(w_1, y^i)$ & $\mathrm{InS}(w_2, y^j)$ & $\mathrm{L}(y^i, l_i)$ & $\mathrm{SWC}(y^i, s_i)$ & $\mathrm{L}(y^j, l_j)$ & $\mathrm{SWC}(y^j, s_j)$ \\
& T5 & 99.6  & 99.5 & 28.8 & 15.2 & 25.3 & 13.7 \\ 
& \mf & 100   & 100 & 56.4 & 28.6 & 55.7 & 27.3 \\
\bottomrule
\end{tabularx}
\label{rocexperimentsclause}
\end{table}

\clearpage

\subsection{Implementation Details For Each Evaluation Task}
In this section, we will introduce the implementation details of all our evaluation tasks. In our experiments, we use three different pre-trained language model, T5-base, T5-Large and MBart-Large. We use the implementation of \emph{huggingface transformers}~\footnote{\url{https://github.com/huggingface/transformers}}. We modify their decoder models to integrate our state matrix and use their provided model weights in our experiment. We only additional introduce the \emph{State Matrix Encoder}. It is a one-layer transformer encoder. Its hidden size equals to the dimension of each head in the pre-trained transformer-based langauge models. The size of its FFN layer is 256. The number of its heads is 4.

\paragraph{Controllable ROCStories Generation}
We first use RAKE algorithm (implemented by \url{https://github.com/csurfer/rake-nltk}) to extract storyline (i.e., key words and phrases) from the ground-truth stories. In the ROCStories dataset, each story has 5 sentences. For extracted storylines, we can easily find their original sentence index and ordering. We can also extract total length and stop word counts from each sentence in the ground-truth stories. We use these information to construct the training rules. For rules with only logic $\land$, we simply use these extracted ground-truth information as the predicate logic constraint. For rules with logic $\lor$ (e.g., $\mathrm{R}_2$ and $\mathrm{R}_3$ in A.4), we create all cases with equal proportion in the training data. For example, for clause $\mathrm{Len}(y^i, l_i) \lor \mathrm{StopWordCount}(y^i, s_i)$, we create 33\% of the training data only satisfy $\mathrm{Len}(y^i, l_i)$, 33\% of the training only satisfy $\mathrm{StopWordCount}(y^i, s_i)$ and the remaining training data satisfy both of them. We can assign fake value for $l_i$ or $s_i$ for the above data argumentation. To improve the generalization of our pre-trained model, we freeze the parameters in the Self-Attention module and Feed-Forward Layers in each layer of the T5 decoder. This parameters freezing technology is applied to both T5 baseline models and the \mf models in all of our experiments. We use constant learning rate $5e^{-5}$ and batch size 32 for this experiment. 

\paragraph{Commonsense Generation} 
In the Commonsense Generation task, we first use NLTK toolkit to expand each input concept with all of its possible inflected forms, including plurals and different tenses. We further search the mention position of each input concept, including all of its inflected forms, on its corresponding ground-truth references. With this mention position, we can construct the state matrix shown in Figure~\ref{commonsenmf} by putting Status 2 after this mention position and Status 0 before this mention position. We use the same model and training setup in the  Controllable ROCStories Generation task. We use constant learning rate $5e^{-5}$ and batch size 48 for this experiment. 

\paragraph{Document-level Machine Translation}
In the document-level Machine Translation, we split each documents into 2 - 4 trucks. Following the fine-tune setup in the original MBart paper, we use learning rate $3e^{-5}$. But we use batch size 8 and total training step 80k for our experiment.

\end{document}